\documentclass[12pt]{article}
\usepackage{amsmath}
\usepackage{times}
\usepackage{graphicx}
\usepackage{color}
\usepackage{multirow}
\usepackage[authoryear]{natbib}
\usepackage{rotating}
\usepackage{bbm}
\usepackage{latexsym}

\newcommand{\captionfonts}{\normalsize}

\makeatletter  
\long\def\@makecaption#1#2{%
  \vskip\abovecaptionskip
  \sbox\@tempboxa{{\captionfonts #1: #2}}%
  \ifdim \wd\@tempboxa >\hsize
    {\captionfonts #1: #2\par}
  \else
    \hbox to\hsize{\hfil\box\@tempboxa\hfil}%
  \fi
  \vskip\belowcaptionskip}
\makeatother   

\usepackage{hyperref}

\usepackage{latexsym,amssymb,lastpage}
\usepackage{graphicx,amsfonts}
\usepackage{mathabx}

\usepackage{times,mathptmx,bm,amsmath}
\usepackage{xcolor}

\usepackage{amsthm}
\usepackage{lipsum}
\usepackage{graphicx}
\usepackage{epstopdf}
\usepackage{algorithmic}
\ifpdf
  \DeclareGraphicsExtensions{.eps,.pdf,.png,.jpg}
\else
  \DeclareGraphicsExtensions{.eps}
\fi

\usepackage{enumitem}

\usepackage{amsopn}

\DeclareMathOperator{\sign}{sign}

\begin{document}


\newcommand{\creflastconjunction}{, and~}


\makeatletter
\newcommand*{\addFileDependency}[1]{
  \typeout{(#1)}
  \@addtofilelist{#1}
  \IfFileExists{#1}{}{\typeout{No file #1.}}
}
\makeatother

\newcommand*{\myexternaldocument}[1]{%
    \externaldocument{#1}%
    \addFileDependency{#1.tex}%
    \addFileDependency{#1.aux}%
}


\newcommand{\A}{\mathsf{A}}
\newcommand{\B}{\mathsf{B}}
\newcommand{\C}{\mathsf{C}}
\newcommand{\D}{\mathsf{D}}
\newcommand{\E}{\mathsf{E}}
\newcommand{\G}{\mathsf{G}}
\renewcommand{\S}{\mathsf{S}}
\newcommand{\LL}{\mathsf{L}}
\newcommand{\M}{\mathsf{M}}
\newcommand{\X}{\mathsf{X}}
\newcommand{\Y}{\mathsf{Y}}
\newcommand{\Z}{\mathsf{Z}}
\newcommand{\x}{\mathsf{x}}
\newcommand{\y}{\mathsf{y}}
\newcommand{\eight}{\mathsf{8}}
\newcommand{\nine}{\mathsf{9}}

\newenvironment{pf}[1][Proof]{\noindent\textit{#1. } }{\hfill$\square$}
  \newtheorem{thm}{Theorem}

\newcommand{\eqdist}{\,{\buildrel d \over =}\,}
\newcommand{\al}{\boldsymbol{\alpha}}
\newcommand{\be}{\boldsymbol{\beta}}

\newcommand{\new}[1]{{#1}}
\newcommand{\simo}[1]{{\color{orange}#1}}
\newcommand{\simoComm}[1]{{\color{purple}#1}}
\newcommand{\paulComm}[1]{{\color{brown}#1}}





\hspace{13.9cm}1

\ \vspace{20mm}\\

\noindent{\LARGE Invariance, encodings, and generalization:\\ learning identity effects with neural networks}

\ \\
{\bf \large S. Brugiapaglia$^{\displaystyle 1}$, M. Liu$^{\displaystyle 1}$, P. Tupper$^{\displaystyle 2}$}\\
{$^{\displaystyle 1}$Department of Mathematics and Statistics, Concordia University, Montr\'eal, QC 
  }\\
{$^{\displaystyle 2}$Department of Mathematics, Simon Fraser University, Burnaby, BC 
  }\\
%


\thispagestyle{empty}
\markboth{Learning Identity Effects}{S. Brugiapaglia, M. Liu, and P. Tupper}

\ \vspace{-0mm}\\
%
\begin{center} {\bf Abstract} \end{center}

Often in language and other areas of cognition, whether two components of an object are identical or not determines if it is well formed. We call such constraints \textit{identity effects}. When developing a system to learn well-formedness from examples,
it is easy enough to build in an identify effect. But can identity effects be learned from the data without explicit guidance? We provide a framework in which we can rigorously prove that algorithms satisfying simple criteria cannot make the correct inference. We then show that a broad class of learning algorithms including deep feedforward neural networks trained via gradient-based algorithms (such as stochastic gradient descent or the Adam method) satisfy our criteria, dependent on the encoding of inputs. In some broader circumstances we are able to provide adversarial examples that the network necessarily classifies incorrectly.   Finally, we demonstrate our theory with computational experiments in which we explore the effect of different input encodings on the ability of algorithms to generalize to novel inputs. \new{This allows us to show similar effects to those predicted by theory for more realistic methods that violate some of the conditions of our theoretical results.}


\section{Introduction}

Imagine that subjects in an experiment are told that the words $\A\A$, $\G\G$, $\LL\LL$, and $\M\M$ are good, and the words $\A\G$, $\LL\M$, $\G\LL$, and $\M\A$ are bad. If they are then asked whether $\Y\Y$ and $\Y\Z$ are good or bad, most will immediately say that $\Y\Y$ is good and $\Y\Z$ is bad. Humans will immediately note that the difference between the two sets of words is that the two letters are identical in the good words, and different in the second. The fact that $\Y$ and $\Z$ do not appear in the training data does not prevent them from making this judgement with novel words.

 However, many machine learning algorithms would not make this same inference given the training set. Depending on how inputs are provided to the algorithm and the training procedure used, the algorithm may conclude that since neither $\Y$ nor $\Z$ appears in the training data,  it is impossible to distinguish two inputs containing them.

The ability or inability of neural networks to generalize learning outside of the training set has been controversial for many years. \cite{marcusbook} has made strong claims in support of the inability of neural networks and other algorithms that do not instantiate variables to truly learn identity effects and other algebraic rules. The explosion of interest in deep neural networks since that book has not truly changed the landscape of the disagreement; see \cite{marcus2019,debateWebsite} for a more recent discussion.
Here we hope to shed some light on the controversy by considering a single instance of an algebraic rule, specifically an identity effect, and providing a rigorous framework in which the ability of an algorithm  to generalize it outside the training set can be studied. 

\new{
The idea of an identify effect comes from linguistics, see e.g.\ \cite{benua1995identity,gallagher2013learning}.  Research in linguistics often focuses on questions such as identifying when a given linguistic structure is well formed or not. Examples include understanding whether a sentence is grammatical (syntax) or whether a word consisting of a string of phonemes is a possible word of a language (phonology). An identity effect occurs when whether a structure is well formed depends on two components of a structure being identical. A particularly clear linguistic example is that of \emph{reduplication}: in many languages words are inflected by repeating all or a portion of the word. For example, in Lakota, an adjective takes its plural form by repeating the last syllable (e.g.\ \emph{h\~{a}ska} [tree] becomes \emph{h\~{a}ska-ska} [trees])  \cite{paschen2021trigger}. In English, we are maybe best familiar with reduplication from the example of \emph{constrastive reduplication} where we might refer to a typical lettuce salad as a  ``salad salad" in order to distinguish it from a (less typical) fruit salad \cite{ghomeshi2004contrastive}. The key point is that linguistic competence with such constructions and others in phonology involves being able to assess whether two items are identical. When an English speaker hears the phrase ``salad salad", to understand it as an instance of contrastive reduplication, the listener must perceive the two uttered words as instances of the same word ``salad", despite any minor phonetic differences in the enunciations.}

\new{Rather than tackling a formalization of identity effects in the linguistic context, we consider an idealization of it that captures the fundamental difficulty of the example of two-letter words we opened with. We take an identify effect task to be one where a learner is presented with two objects (encoded in some way, such as a vector of real values) and must determine whether these two objects are identical in some relevant sense. Sometimes this will mean giving a score of 1 to a pair of objects that are actually identical (their encodings are exactly the same) and 0 otherwise, or it may mean that the learner must determine if they are representatives of the same class of objects. In either case, we want to determine which learners can, from a data set of pairs of identical and nonidentical objects, with the correct score given, generalize to make the same judgements with different pairs of objects, including ones not in the training set.}


\new{The difficulty of learning identity effects is just one application of our theory of learning and generalization under transformations.}
In our framework, we consider mappings that transform the set of inputs, and consider whether particular learning algorithms are invariant to these transformations, in a sense which we will define. We show that if both the learning algorithm and the training set are invariant to a transformation, then the predictor learned by the learning algorithm is also invariant to the transformation, meaning that it will assess inputs before and after transformation as equally well formed.
 We apply our results to the learning of identity effects. We define a mapping that, in the example above, leaves the training data unchanged, but swaps the inputs $\Y\Y$ and $\Y\Z$, and so any learning algorithm that is invariant to that map cannot distinguish between these two inputs.
We then show that a broad class of algorithms, including deep feedforward neural networks trained via stochastic gradient descent, are invariant to the same map for some commonly used encodings. Furthermore, for other encodings we show how to create an adversial example to ``trick'' the network into giving the wrong judgment for an input.
Finally, we show with computational experiments how this dependence on encoding plays out in practice. In our example we will see that one-hot encoding (also known as localist encoding) leads to a learner that is unable to generalize outside the training set, whereas distributed encoding allows partial generalization outside the training set. 





In Section~\ref{sec:mainResult} we provide the framework for our theory and prove the main result: Rating Impossibility for Invariant Learners. 
In Section~\ref{sec:identity} we apply our theory to the case of identity effects of the type in our motivating example.
We then  show that the conditions of the theorem comprising our main result are satisfied for a broad class of algorithms including neural networks trained via stochastic gradient descent and with appropriate encodings. For other encodings we show how to create adversarial examples for which the network will give the wrong answer even for inputs whose two components are identical.
Then in Section~\ref{sec:experiments} we demonstrate the theory with numerical experiments. We examine the ability of learning algorithms to generalize the identity effect with the task in the opening of our paper, first with pairs of letter and abstract encodings, and then with pairs of numbers where each number is represented by distinct hand-drawn digits from the MNIST data set of \cite{lecun2010mnist}.
\new{Our numerical experiments show that in many cases, some practical learning algorithms, though not covered explicitly by our theory, show many of the same obstacles that we established earlier for theoretically simpler algorithms.}


\section{Main results} \label{sec:mainResult}


Suppose we are training an algorithm to assign \new{real number} ratings to inputs.  Often the ratings will just be 0 or 1, like in the case of a binary classifier, but they also can \new{also take values in an interval}.
Let $W$ be the set of all possible inputs $w$. \new{There is no constraint on $W$, though we can imagine $W$ to be $\mathbb{R}^d$ or the set of all finite strings composed from a given set of letters.}
Our learning algorithm is trained on a \new{data set} $D$ consisting of a \new{finite} list of input-output pairs $(w,r)$ where $w \in W$ and $r \in \mathbb{R}$. Let $\mathcal{D}$ be the set of all possible data sets with inputs from $W$. (In the motivating example introduced in the opening paragraph, $W$ is the set of all possible two-letter words.)

Typically, in machine learning there is a training algorithm (such as stochastic gradient descent) which takes as input a training data set $D$ and outputs a set of parameters $\new{\Theta}$, defining a model $f$. We formalize this with a map $\mathcal{A}$ as 
\[
\new{\Theta}= \mathcal{A}(D).
\]
Note that the training algorithm might involve randomized  operations, such as random parameter initialization; in this case, the set of parameters $\new{\Theta}$ is a random variable. For the moment, let us assume  $\mathcal{A}$ to be deterministic. When we want to give a rating to a novel input $w$, we plug it into our model $f$ using the parameters $\new{\Theta}$, i.e.
\[
r= f(\new{\Theta},w).
\] 
In the case of artificial neural networks, this operation  corresponds to a forward propagation of $w$ through the trained network.

Though in practice determining $\new{\Theta}$ is done separately from computing the rating of $w$ (especially since one usually wants multiple $w$ to be evaluated), for our purposes we can combine them into one function.
We define the  learning algorithm as a map $L \colon \mathcal{D} \times W \rightarrow \mathbb{R}$ given by
\[
L(D,w) = f(\mathcal{A}(D), w).
\]

We want to be able to show that a given algorithm is not able to distinguish between two inputs not in $D$. More formally, we want our conclusion to be of the form
\[
L(D,w_1) = L(D,w_2),
\]
for two inputs $w_1 \new{\neq} w_2$ in $W$, but not in $D$, when $L$ and $D$ have some particular structure.

The relation between $w_1$ and $w_2$ will be defined with the help of a function
 $\tau \colon W \rightarrow W$ that takes $w \in W$ and gives $\tau(w) \in W$. 
For example, if $W$ is a set of words, $\tau$ might reverse the order of the letters. If $W$ is a set of images, $\tau$ might perform a mirror reflection. In the case of a data set $D$, we define $\tau(D)$ as the data set obtained by replacing every instance of $(w,r)$ in $D$ with $(\tau(w),r)$.

Our main result follows.\ 

\begin{thm}[Rating impossibility for invariant learners] \label{thm:main}

\new{Consider a data set $D\in\mathcal{D}$ and a transformation $\tau:W \to W$ such that 
\begin{enumerate}
\item $\tau(D) = D$ (invariance of the data). 
\end{enumerate}
Then, for any learning algorithm $L: \mathcal{D} \times W \to \mathbb{R}$ and any input $w \in W$ such that
\begin{enumerate}[resume]
\item $L(\tau(D), \tau(w)) = L(D,w)$ (invariance of the  algorithm),
\end{enumerate}
we have $L(D,\tau(w)) = L(D,w)$.}

\end{thm}

\begin{pf}
$$
L(D,\tau(w)) = L(\tau(D), \tau(w)) = L(D,w).
$$
\end{pf}

\new{The first condition, \textit{invariance of the data}, we expect to hold only for certain particular data sets, and, in particular, the richer the data set, the fewer transformations $\tau$ it will be invariant to. The second condition in the theorem, \textit{invariance of the algorithm}, we will show to be true of some learning procedures for all $D$ and $w$, though the result only requires it for the $D$ and $w$ of interest.}
  Under these two conditions, the theorem states that the algorithm will not  give different ratings to $w$ and $\tau(w)$ when trained on $D$.

Here is a simple example of how this theorem works. Suppose $W$ consists of two-letter words and $\tau$ is a transformation that reverses the order of the two letters.
Suppose $L$ is a learning algorithm that is invariant to $\tau$ \new{for $D$ and all $w \in W$}, which is a fairly reasonable assumption, unless we explicitly build into our algorithm reason to treat either letter differently.
Suppose $D$ is a training set where all the words in it are just the same letter twice, so that $\tau(D) = D$. Then the theorem states that the learning algorithm trained on $D$ will give the same result for $w$ and $\tau(w)$ for all words $w$. So the algorithm will give the same rating to $xy$  and $yx$ for all letters $x$ and $y$. This is not surprising: if the algorithm has no information about words $xy$ where $x \neq y$, then why would it treat $xy$ and $yx$ differently?

Up until now, we have let our set of inputs $W$ be any set of objects. But in practice, our inputs will always be encoded as vectors. We use $w$ to denote both the input and its encoded vector. \new{In the latter case, we assume $w \in \mathbb{R}^d$, for some $d \in \mathbb{N}$.} We will also consider maps $\tau$ that are implemented by linear transformations when working with encoded vectors. We denote the linear transformation that implements $\tau$ by \new{$w \mapsto \mathcal{T} w$, for some matrix $\mathcal{T}\in\mathbb{R}^{d \times d}$}.  As an example, consider the situation in the previous paragraph. We assume that each letter in the alphabet has some encoding as a vector of length $n$ and each two-letter word can be encoded by concatenating the two corresponding vectors for the letter together to get a vector of length $2n$. Then the map $\tau$ that switches the order of the letter is implemented by a $2n \times 2n$ permutation matrix $\mathcal{T}$ that swaps the first $n$ entries of a vector with the last $n$ entries.

In Section~\ref{sec:identity} we will show how to apply the theorem to identity effects, and in particular to our motivating example.


Using Theorem~\ref{thm:main} requires that we actually establish invariance of our algorithm \new{for a given $D$ and $w$} for the relevant transformation when inputs are encoded in a particular way. Here we establish invariance \new{for some $D$ and $w$} for some classes of transformation $\tau$ and for some popular machine learning frameworks and encodings.
We  assume that our learning algorithm works by using a model for the data in which there are parameters. The parameters are then fit 
by minimizing a loss function on training data.

\subsection{No regularization}

We suppose our model for the data $D=\{(w_i,r_i)\}_{i = 1}^n$ is given by $r =f(B, C  w)$ where  $C$ is a matrix containing the coefficients multiplying $w$ and $B$ incorporates all other parameters including any constant term added to $C w$ (e.g., the first bias vector in the case of artificial neural networks). The key point is that the parameters $C$ and the input $w$ only enter into the model through $C w$. \new{Note that there is a slight abuse of notation here since we assume that $f(\Theta, w) = f(B,C w)$, where $\Theta = (B,C)$.}

This at first might seem restrictive, but in fact most neural network models use this structure: input vectors are multiplied by a matrix of parameters before being processed further. For example, suppose we are training a three-layer feedforward neural network whose output $r$ is given by
\[
r=\sigma_3( W_3 \, \sigma_2( W_2 \, \sigma_1( W_1 w +b_1) + b_2) +b_3),
\]
where $W_1, W_2, W_3$ are weight matrices, $b_1$, $b_2$, $b_3$ are bias vectors, and $\sigma_1, \sigma_2, \sigma_3$ are nonlinear activations (e.g., ReLU or sigmoid functions). 
In this case, we can let $C = W_1$ and $B=(W_2, W_3, b_1, b_2, b_3)$ to show that it fits into the required form.

Now suppose we select $B$ and $C$ by optimizing some loss function
\begin{equation}
\label{eq:loss_function}
F(B,C) = \mathcal{L}( f(B, C  w_i),r_i, i=1\ldots n),
\end{equation}
\new{so that $B$ and $C$ implicitly depend on $D$.}
For example, $F(B,C) = \sum_{i = 1}^n (r_i - f(B,Cw_i))^2$ when the mean squared error is used as a loss function. Moreover, \new{we} assume  that the loss function is minimized by a unique set of values \new{for all $D$}.
In the following theorem, under these conditions we obtain invariance of the algorithm (condition 2. of Theorem~\ref{thm:main}) for any transformation $\tau$ that is linear and invertible.

\begin{thm} \label{thm:noRegularization}
\new{Consider a loss function of the form \eqref{eq:loss_function} that admits, for any data set $D$, a unique minimizer $(\hat{B},\hat{C})$ (implicitly depending on $D$).
Suppose that a learning algorithm $L$ evaluates inputs according to 
\[
L(D,w)= f(\hat{B},\hat{C}w).
\]}
Then, for any $D$ and $w$, $L$ is invariant to any $\tau$ that is a linear invertible transformation: 
$$
L(\tau(D), \tau(w))=L(D,w).
$$
\end{thm}

\begin{pf}
Since $\tau$ is linear and invertible it can be expressed as $\tau(w) = \mathcal{T} w$, for some invertible matrix $\mathcal{T}$. If we apply $\mathcal{T}$ to the words $w_i$ in the data set and perform optimization again, we get new parameters $B'$ and $C'$.
But note that $C' (\mathcal{T} w_i) = (C' \mathcal{T}) w_i$. So the optimum is obtained by letting
\[
 C' \mathcal{T} = \hat{C},
\]
or $C' = \hat{C} \mathcal{T}^{-1}$, and $B' = \hat{B}$.
We then obtain
\[
L(\tau(D), \tau(w)) = f(B',C' \mathcal{T} w)= f(B,\hat{C} w) = L(D,w),
\]
as required. 
\end{pf}

\new{The assumption that there is a unique set of parameters that minimizes the loss function for every data set $D$ is of course very strong, and is unlikely to hold in practice. It holds for simple linear regression with mean square loss function, but is unlikely to hold for more complicated models (due to nonuniqueness of parameter values) and for other loss functions, such as the cross-entropy loss function. In the case of cross-entropy loss function, without regularization, arbitrarily large parameter values attain increasingly small values of loss, and there are no parameter values that attain a minimum. In practice, effective parameter values are obtained either by regularization (see Subsection \ref{subsec:reg}) or by early termination of the optimization algorithm (see Subsection~\ref{sec:SGD_theory}). We offer this result, limited though it may be in application, because it contains, in simpler form, some of the ideas that will appear in later results.}

\subsection{Regularization} \label{subsec:reg}

So far we have considered a loss function where the parameters $C$ that we are fitting only enter through the model $f$ in the form $C w_i$. But, more generally, we may consider the sum of a loss function and a regularization term: 
\begin{equation} \label{eq:withRegularization}
F(B,C) = \mathcal{L}( f(B, C  w_i),r_i, i=1\ldots n)+ \lambda \mathcal{R}(B,C),
\end{equation}
where $\lambda \geq 0$ is a tuning parameter, and  suppose $B$ and $C$ are obtained by minimizing this objective function.

\begin{thm} \label{thm:withRegularization}
\new{Consider a regularized loss function of the form \eqref{eq:withRegularization} that admits, for any data set $D$, a unique minimizer $(\hat{B},\hat{C})$ (implicitly depending on $D$).}
Suppose that a learning algorithm $L$ evaluates inputs according to 
\[
L(D,w)= f(\hat{B},\hat{C}w).
\]
Suppose $\tau$ is a linear invertible transformation with $\tau (w)= \mathcal{T} w$ for some matrix $\mathcal{T}$, and that the regularization term satisfies 
$\mathcal{R}(B, C \mathcal{T}) =\mathcal{R}(B,C)$. 
Then, for any $D$ and $w$, $L$ is invariant to  $\tau$ : 
$$
L(\tau(D), \tau(w))=L(D,w).
$$
\end{thm}

\begin{pf}
The proof goes through exactly as in Theorem~\ref{thm:noRegularization}, because of the condition $\mathcal{R}(B, C \mathcal{T}) $ $=\mathcal{R}(B,C)$.
\end{pf}

This invites the question: for a given choice of regularization, which linear transformations $\tau$ will satisfy the conditions of the theorem? The only condition involving the regularization term  is $\mathcal{R}(B, C \mathcal{T}) =\mathcal{R}(B,C)$. So, if $\mathcal{R}$ has the form
\[
\mathcal{R}(B,C)= \mathcal{R}_1(B) + \| C\|^2_F,
\]
where $\| \cdot \|_F$ is the Frobenius norm (also known as $\ell^2$ regularization) and where $\mathcal{R}_1(B)$ is a generic regularization term for $B$, then any  transformation $\tau$ represented by an orthogonal matrix $\mathcal{T}$ will lead to a learning algorithm that is invariant to $\tau$. In fact, $\|C\mathcal{T}\|_F = \|C\|_F$ for any orthogonal matrix $\mathcal{T}$.
If we use $\ell^1$ regularization for $C$, corresponding to 
$$
\mathcal{R}(B,C) = \mathcal{R}_1(B) + \|C\|_1,
$$
where $\|\cdot\|_1$ is the sum of the absolute values of the entries of $C$, the algorithm will not be invariant to all orthogonal transformations. However, it will be invariant to transformations $\tau$ that are implemented by a signed permutation matrix $\mathcal{T}$. As we will discuss in Section~\ref{sec:appl_IE}, this will be the case in our motivating example with one-hot encoding.

\subsection{Stochasticity and gradient-based training} 
\label{sec:SGD_theory}

Up to this point, we have assumed that our classifier is trained deterministically by finding the unique global minimizer of an objective function. In practice, an iterative procedure is used to find values of the parameters that make the loss function small, but even a local minimum may not be obtained.
For neural networks, which are our focus here, a standard training method is stochastic gradient descent (SGD) (see, e.g., \citet[Chapter 8]{goodfellow2016deep}). Parameters are determined by randomly or deterministically generating initial values and then using gradient descent to find values that sufficiently minimize the loss function. Rather than the gradient of the whole loss function, gradients are computed based on a randomly chosen batch of training examples at each iteration.
So stochasticity enters both in the initialization of parameters and in the subset of the data that is used for training in each step of the algorithm. Here we show that our results of the previous subsections extend to SGD with these extra considerations; in the Supplemental Information we consider the case of the Adam method (see \cite{kingma2014adam}).


\new{
In what follows our parameter values, and the output of a learning algorithm using those parameter values, will be random variables, taking values in a vector space. The appropriate notion of equivalence between two such random variable for our purposes (which may be defined on different probability spaces) is equality in distribution \cite{billingsley2008probability}. To review, two random variables $X$ and $Y$ taking values in $\mathbb{R}^k$ are equal in distribution (denoted $X \eqdist Y$) if for all $x \in \mathbb{R}^k$
\[
\mbox{Prob}(X_i \leq x_i \colon i=1,\ldots,k) = \mbox{Prob}(Y_i \leq x_i \colon i=1,\ldots,k).
\]
For any function $g \colon \mathbb{R}^k \rightarrow \mathbb{R}$, when $X \eqdist Y$, we have $\mathbb{E}g(X)= \mathbb{E}g(Y)$, whenever both sides are defined. 
This means that if the output of two learning procedures is equal in distribution, then the expected  error on a new data point is also equal.
}

Let $D$ be our complete data set with entries $(w,r)$ and suppose our goal is to find parameters $B,C$ that minimize, \new{for some fixed $\lambda \geq 0$},
\[
F(B,C)= \mathcal{L}(f(B,Cw),r | (w,r) \in D) + \lambda(\mathcal{R}_1(B) +  \|C\|^2_F),
\]
so that we can use $L(D,w)=f(B,Cw)$ as our classifier. In order to apply SGD, we will assume the function $F$ to be differentiable with respect to $B$ and $C$. \new{Since $\lambda \geq 0$, our discussion includes the cases of regularization and no regularization}. For subsets $D_i$ of the data $D$ let us define $F_{D_i}$ to be $F$ but where the loss function is computed only with data in $D_i$.
In SGD we randomly initialize the parameters $B_0$ and $C_0$, and then take a series of steps 
$$
B_{i+1} = B_{i} - \theta_{i} \frac{\partial F_{D_i}}{\partial B} (B_i, C_i), \ \ \ 
C_{i+1} = C_{i} - \theta_{i} \frac{\partial F_{D_i}}{\partial C} (B_i, C_i),
$$
for $i = 0,1,\ldots,k-1$ 
where we have  a predetermined sequence of step sizes $\{\theta_i\}_{i = 1}^{k-1}$,
and $D_i$ are a randomly selected subsets (usually referred to as ``batches'' or ``minibatches'') of the full data set $D$ for each $i$. We assume that the $D_i$ are selected \new{either deterministically according to some predetermined schedule or randomly at each time step but in either case, independently of all previous values of $(B_i, C_i)$.} For each $i$, $(B_i,C_i)$ are random variables, and therefore the output of the learning algorithm $L(D,w)=f(B_k,C_k w) $ is a random variable.
We want to show for certain transformations $\tau$ that $L(D,w)$ \new{has} the same distribution as $L(\tau(D),\tau(w))$, i.e.\ $L(D,w) \eqdist L(\tau(D),\tau(w))$.

We randomly initialize the parameters $C$ as $C = C_0$, such that $C_0$ and $C_0\mathcal{T}$ have the same distribution. This happens, for example, when the entries of $C_0$ are identically and independently distributed according to a normal distribution $\mathcal{N}(0,\sigma^2)$. (Note that this scenario includes the deterministic initialization $C_0 = 0$, corresponding to $\mathcal{N}(0,0)$). We also  initialize $B=B_0$ in some randomized or deterministic way independently of $C_0$.

Now, what happens if we apply the \textit{same} training strategy using the transformed data set $\tau(D)$? We denote the generated parameter sequence with this training data $\{(B'_i,C'_i)\}_{i = 1}^k$.
In the proof of the following theorem we show that the sequence $(B'_i,C'_i \mathcal{T})$ has the same distribution as $(B_i,C_i)$ for all $i$. Then, if we use $(B_k,C_k)$ as the parameters in our model we obtain
$$
L(\tau(D),\tau(w)) = f(B'_k,C'_k \mathcal{T} w ), 
$$
which has the same distribution as $f(B_k,C_k w) = L(D,w)$, establishing invariance of the learning algorithm to $\tau$.
The full statement of our results is as follows.

\begin{thm} \label{thm:SGD}
Let $\tau$ be a linear transformation with orthogonal matrix $\mathcal{T}$.
Suppose SGD, as described above, is used
to determine parameters $(B_k,C_k)$ with the objective function
\[
F(B,C)= \mathcal{L}(f(B,Cw_i),r_i, i=1,\ldots,n) + \lambda(\mathcal{R}_1(B) +  \|C\|^2_F),
\]
for some $\lambda \geq 0$ and assume  $F$ to be differentiable with respect to $B$ and $C$. Suppose the random initialization of the parameters $B$ and $C$ to be independent and that the initial distribution of $C$ is invariant with respect to right-multiplication by $\mathcal{T}$.
 
Then, the learner $L$ defined by $L(D,w)=f(B_k,C_k w)$ satisfies $L(D,w) \eqdist L(\tau(D),\tau(w))$. 
\end{thm} 
\begin{proof}
Let $(B'_0,C'_0) \eqdist (B_0,C_0)$ and let $(B'_i,C'_i)$, $i=1,\ldots,k$ be the sequence of parameters generated by SGD with the transformed data $\tau(D)$. Each step of the algorithm uses a transformed subset of the data $\tau(D_i)$.  By hypothesis, $(B_0,C_0)\eqdist (B'_0,C'_0 \mathcal{T})$. We will show that $(B_i,C_i) \eqdist (B'_i,C'_i \mathcal{T})$ for all $i$. 
Using induction, let us suppose they are identical for a given $i$, and then show  they are also identical for $i+1$.

First let's note that because $F_{D_i}$ only depends on the input words $w$ and $C$ through expressions of the form $C w$ and thanks to the form of the regularization term $\mathcal{R}_1(B) + \|C\|_F$ we have that
$F_{\tau(D_i)}(B,C)= F_{D_i}(B,C\mathcal{T})$. So 
\begin{eqnarray*}
\frac{\partial F_{\tau(D_i)}}{\partial B} (B,C) & = & \frac{\partial F_{D_i}}{\partial B} (B,C \mathcal{T}), \\
\frac{\partial F_{\tau(D_i)}}{\partial C} (B,C) & = & \frac{\partial F_{D_i}}{\partial C} (B,C \mathcal{T}) \mathcal{T}^T. 
\end{eqnarray*}
With these results we have 
\begin{eqnarray*}
B'_{i+1} &=& B'_{i} - \theta_{i} \frac{\partial F_{\tau(D_i)}}{\partial B} (B'_i, C'_i), \\
& =& B'_{i} - \theta_{i} \frac{\partial F_{D_i}}{\partial B} (B'_i, C'_i \mathcal{T}), \\
& \eqdist & B_{i} - \theta_{i} \frac{\partial F_{D_i}}{\partial B} (B_i, C_i) =B_{i+1},
\end{eqnarray*}
where we used  the inductive hypothesis in the last line. 

For $C'_{i+1}$ we have
\begin{eqnarray*}
C'_{i+1} &=& C'_{i} - \theta_{i} \frac{\partial F_{\tau(D_i)}}{\partial C} (B'_i, C'_i), \\
& =& C'_{i} - \theta_{i} \frac{\partial F_{D_i}}{\partial C} (B'_i, C'_i \mathcal{T}) \mathcal{T}^{T}, \\
& \eqdist & C_{i} \mathcal{T}^{-1} - \theta_{i} \frac{\partial F_{D_i}}{\partial C} (B_i, C_i) \mathcal{T}^T =C_{i+1} \mathcal{T}^{-1},
\end{eqnarray*}
where we have used the fact that $\mathcal{T}$ is an orthogonal matrix. This establishes $C_{i+1} \eqdist C'_{i+1} \mathcal{T}$.

Now we have that $(B_i,C_i) \eqdist (B'_i,C'_i \mathcal{T})$ and so
\[
L(\tau D,\tau w)= f(B_k',C'_k \mathcal{T} w) \eqdist f(B_k,C_k w).
\]
\end{proof}


\subsection{Recurrent neural networks}

We now illustrate how to apply our theory to the case of Recurrent Neural Networks (RNNs)  \citep{rumelhart1986learning}. This is motivated by the fact that a special type of RNNs, namely Long-Short Term Memory (LSTM) networks, have been recently employed in the context of learning reduplication in \cite{prickett2018seq2seq, prickett2019learning}. Note also that numerical results for LSTMs in the contetx of learning identity effects will be illustrated in Section~\ref{sec:experiments}. RNNs (and, in particular, LSTMs) are designed to deal with inputs that possess a sequential structure. 
From a general viewpoint, given an input sequence $w = (w^{(t)})_{t=1}^T$ an RNN computes a sequence of hidden units $h = (h^{(t)})_{t = 1}^{T}$ by means of a recurrent relation of the form $h^{(t)} = g(w^{(t)},h^{(t-1)}; \Theta)$ for some function $g$, trainable parameters $\Theta$, and for some given initial value $h^{(0)}$. The key aspect is that the same $g$ is applied to all inputs $w^{(t)}$ forming the input sequence. Note that this recurrent relation can be ``unfolded'' in order to write $h^{(t)}$ as a function of $w^{(1)},\ldots,w^{(t)}$ without using recurrence. The sequence $h$ is then further processed to produce the network output. We refer to \citet[Chapter 10]{goodfellow2016deep} for more technical details on RNNs and LSTMs. 

Here, we will assume the input sequence to have length two and denote it by $w = (u,v)$. In other words, the input space is a Cartesian product $W = U \times U$\new{, for some set $U$. There is no constraint on $U$, but we can imagine $U$ to be $\mathbb{R}^d$ or a given set of letters}. This is natural in the context of identity effects since the task is to learn whether two elements $u$ and $v$ of a sequence $w=(u,v)$ are identical or not. We consider learners of the form 
$$
L(D,w) = f(B, C u, Cv), \quad w = (u,v),
$$
where $B,C$, are trained parameters. This includes a large family of RNNs and, in particular, LSTMs (see, e.g., \citet[Section 10.10.1]{goodfellow2016deep}). Note that the key difference with respect to a standard feedforward neural network is that $u$ and $v$ are multiplied by the same weights $C$ because of the recurrent structure of the network. Using block matrix notation and identifying $u$ and $v$ with their encoding vectors, we can write
$$
L(D,w) = f\left(B, \begin{bmatrix}C & 0 \\ 0 & C\end{bmatrix}\begin{bmatrix}u \\v\end{bmatrix}\right).
$$
This shows that the learner is still of the form $L(D,w) = f(B,C'w)$, analogously to the previous subsection. However, in the RNN case $C'$ is constrained to have a block diagonal structure with identical blocks on the main diagonal.  In this framework, we are able to prove the following invariance result, with some additional constraints on the transformation $\tau$. We are not able to obtain results for regularization on both $B$ and $C$, though our results apply to common practice, since LSTM training is often performed without regularization (see, e.g., \cite{greff2016lstm}). We will discuss the implications of this result for learning identity effects in Section~\ref{sec:appl_IE}.

\begin{thm} \label{thm:RNNs}
Assume the input space to be of the form $W = U \times U$. Let $\tau:W\to W$ be a linear transformation defined by  $\tau(w) = (u, \tau_2(v))$ for any $w = (u,v) \in W$, where $\tau_2: U \to U$ is also linear. Moreover, assume that: 
\begin{itemize}
    \item [(i)] the matrix $\mathcal{T}_2$ associated with the transformation $\tau_2$ is orthogonal and symmetric;
    \item [(ii)] the data set $D=\{((u_i,v_i),r_i)\}_{i=1}^n$ is invariant under the transformation $\tau_2 \otimes \tau_2$, i.e.
\begin{equation}
\label{eq:assumption_RNNs}
(u_i,v_i) = (\tau_2(u_i), \tau_2(v_i)), \quad  i = 1,\ldots,n.
\end{equation}
\end{itemize}
Suppose SGD, as described in Subsection \ref{sec:SGD_theory}, is used to determine parameters $(B_k,C_k)$ with objective function 
\begin{equation}
\label{eq:loss_RNNs}
F(B,C)=  \sum_{i=1}^n \ell(f(B, Cu_i, C v_i) , r_i) 
+ \lambda\mathcal{R}_1(B),
\end{equation}
for some $\lambda \geq 0$, where $\ell$ is a real-valued function and where $\ell$, $f$, and $\mathcal{R}_1$ are differentiable. Suppose the random initialization of the parameters $B$ and $C$ to be independent and that the initial distribution of $C$ is invariant with respect to right-multiplication by $\mathcal{T}_2$.
 
Then, the learner $L$ defined by $L(D,w)=f(B_k,C_ku, C_kv)$, where $w = (u,v)$, satisfies $L(D,w) \eqdist L(\tau(D),\tau(w))$. 
\end{thm} 

\begin{proof}
Given a batch $D_i \subseteq D$, let us denote
$$
F_{D_i}(B,C)=  \sum_{j \in D_i} \ell(f(B, Cu_j, C v_j) , r_j) 
+ \lambda \mathcal{R}_1(B).
$$
The proof is similar to Theorem~\ref{thm:SGD}. However, in this case we need to introduce an \emph{auxiliary objective function}, defined by
$$
\widetilde{F}_{D_i}(B,G,H) 
= \sum_{j \in D_i} \ell(f(B, G u_j, H v_j) , r_j) + \lambda \mathcal{R}_1(B),
$$
Then, $F_{D_i}(B,C) = \widetilde{F}_{D_i}(B,C,C)$ and
\begin{align}
\label{eq:RNN_grad_B}
\frac{\partial F_{D_i}}{\partial B}(B,C)  &= \frac{\partial \widetilde{F}_{D_i}}{\partial B}(B,C,C),\\
\label{eq:RNN_grad_C}
\frac{\partial F_{D_i}}{\partial C}(B,C)  &= \frac{\partial \widetilde{F}_{D_i}}{\partial G}(B,C,C) + \frac{\partial \widetilde{F}_{D_i}}{\partial H}(B,C,C).
\end{align}
Moreover, replacing $D_i$ with its transformed version $\tau(D_i) = \{((u_j,\tau_2(v_j)), r_j)\}_{j\in D_i}$, we see that
$F_{\tau(D_i)}(B,C) = \widetilde{F}_{D_i}(B, C, C \mathcal{T}_2)$. (Note that, as opposed to the proof of Theorem~\ref{thm:SGD}, it is not possible to reformulate $F_{\tau(D_i)}$ in terms of $F_{D_i}$ in this case -- hence the need for an auxiliary objective function). This leads to 
\begin{align}
\label{eq:RNN_grad_tau_B}
\frac{\partial F_{\tau(D_i)}}{\partial B}(B,C)  & = \frac{\partial \widetilde{F}_{D_i}}{\partial B}(B,C, C\mathcal{T}_2),\\
\label{eq:RNN_grad_tau_C}
\frac{\partial F_{\tau(D_i)}}{\partial C}(B,C)  & = \frac{\partial \widetilde{F}_{D_i}}{\partial G}(B,C, C\mathcal{T}_2) + \frac{\partial \widetilde{F}_{D_i}}{\partial H}(B,C, C\mathcal{T}_2) \mathcal{T}_2^T.
\end{align}
Now, denoting $\ell = \ell(f,r)$ and $f = f(B, u,v)$, we have
\begin{align*}
\frac{\partial \widetilde{F}_{D_i}}{\partial G} 
= \sum_{j\in D_i} \frac{\partial \ell}{\partial f} \frac{\partial f}{\partial u} u_j^T, 
\qquad
\frac{\partial \widetilde{F}_{D_i}}{\partial H} 
= \sum_{j \in D_i} \frac{\partial \ell}{\partial f} \frac{\partial f}{\partial v} v_j^T.
\end{align*}
Thanks to the assumption \eqref{eq:assumption_RNNs}, we have $u_j^T \mathcal{T}_2^T = u_j^T$ and $v_j^T \mathcal{T}_2^T = v_j^T$ for all $j \in D_i$. Thus, we obtain 
\begin{equation}
\label{eq:RNN_magic_relation}
\frac{\partial \widetilde{F}_D}{\partial G} \mathcal{T}_2^T = \frac{\partial \widetilde{F}_D}{\partial G},
\qquad \frac{\partial \widetilde{F}_D}{\partial H} \mathcal{T}_2^T = \frac{\partial \widetilde{F}_D}{\partial H}.
\end{equation}

Now, let $(B'_0,C'_0) \eqdist (B_0,C_0)$ and let $(B_i',C_i')$, with $i=1,\ldots,k$ be the sequence generated by SGD, as described in Subsection~\ref{sec:SGD_theory}, applied to the transformed data set $\tau(D)$. By assumption, we have $B_0' \eqdist B_0$ and $C_0 \eqdist C'_0 \eqdist C_0' \mathcal{T}_2$. We will show by induction that $B_0' \eqdist B_0$ and $C_0 \eqdist C'_0 \eqdist C_0' \mathcal{T}_2$ for all indices $i=1,\ldots,k$. 
On the one hand, using \eqref{eq:RNN_grad_B}, \eqref{eq:RNN_grad_tau_B}, and the inductive hypothesis, we have
\begin{align*}
B_{i+1}' 
& = B_i' - \theta_i \frac{\partial F_{\tau(D_i)}}{\partial B}(B_i', C_i')\\
& =  B_i' - \theta_i \frac{\partial \tilde{F}_{D_i}}{\partial B}(B_i', C_i', C_i'\mathcal{T}_2)\\
& \eqdist  B_i - \theta_i \frac{\partial \tilde{F}_{D_i}}{\partial B}(B_i, C_i, C_i\mathcal{T}_2)\\
& = B_i - \theta_i \frac{\partial F_{D_i}}{\partial B}(B_i, C_i)= B_{i+1}.
\end{align*}
On the other hand, using \eqref{eq:RNN_grad_C}, \eqref{eq:RNN_grad_tau_C}, \eqref{eq:RNN_magic_relation} and the inductive hypothesis, we see that
\begin{align*}
C_{i+1}' 
& = C_i' - \theta_i \frac{\partial F_{\tau(D_i)}}{\partial C}(B_i', C_i')\\
& = C_i' - \theta_i \left(
\frac{\partial \tilde{F}_{D_i}}{\partial G}(B_i', C_i', C_i'\mathcal{T}_2) 
+ \frac{\partial \tilde{F}_{D_i}}{\partial H}(B_i', C_i', C_i'\mathcal{T}_2) \mathcal{T}_2^T
\right)\\
& = C_i' - \theta_i \left(
\frac{\partial \tilde{F}_{D_i}}{\partial G}(B_i', C_i', C_i'\mathcal{T}_2) 
+ \frac{\partial \tilde{F}_{D_i}}{\partial H}(B_i', C_i', C_i'\mathcal{T}_2) 
\right)\\
& \eqdist C_i - \theta_i \left(
\frac{\partial \tilde{F}_{D_i}}{\partial G}(B_i, C_i, C_i\mathcal{T}_2) 
+ \frac{\partial \tilde{F}_{D_i}}{\partial H}(B_i, C_i, C_i\mathcal{T}_2) 
\right)\\
& = C_i - \theta_i \frac{\partial F_{\tau(D_i)}}{\partial C}(B_i, C_i) = C_{i+1}.
\end{align*}
Similarly, one also sees that $C_{i+1}'\mathcal{T}_2 \eqdist C_{i+1}$ using \eqref{eq:RNN_grad_C}, \eqref{eq:RNN_grad_tau_C}, \eqref{eq:RNN_magic_relation}, the inductive hypothesis, combined with the symmetry and orthogonality of $\mathcal{T}_2$.

In summary, this shows that 
$$
L(D,w) 
= f(B_k, C_ku, C_k v) 
\eqdist f(B_k', C_k'u, C_k' v)
\eqdist f(B_k', C_k'u, C_k' \mathcal{T}_2 v)
= L(\tau(D), \tau(w)),
$$
and concludes the proof.
\end{proof}

We conclude by observing that loss functions of the form 
$$
\mathcal{L}(((u_i,v_i), r_i), i = 1,\ldots,n) 
= \sum_{i = 1}^n \ell(f(B,C u_i,C v_i), r_i),
$$
such as the one considered in \eqref{eq:loss_RNNs}, are widely used in practice. These include, for example, the mean squared error loss, where $\ell(f,r) = |f-r|^2$, and the cross-entropy loss, where $\ell(f,r) = -r\log(f)- (1-r)\log(1-f)$. 


\section{Application to Identity Effects} \label{sec:identity}


\subsection{Impossibility of correct ratings for some encodings}

\label{sec:appl_IE}

We now discuss how to apply our results to our actual motivating example, i.e.\ learning an identity effect.
 Again, suppose words in $W$ consist of ordered pairs of capital letters from the English alphabet.
 Suppose our training set $D$ consists of, as in our opening paragraph, a collection of two-letter words none of which contain the letters $\Y$ or $\Z$.  The ratings of the words in $D$ are $1$ if the two letters match and $0$ if they don't. \new{We want to see if our learner can generalize this pattern correctly to words that did not appear in the training set, in particular to words containing just $\Y$ and $\Z$.}
%
To apply Theorem~\ref{thm:main}, let $\tau$ be defined by
\begin{equation}
\label{eq:def_tau_IE}
\tau(x\Y)=x\Z, \ \ \ \tau(x\Z)=x\Y, \ \ \ \tau(xy)=xy, 
\end{equation} 
for all letters $x$ and $y$ with $y \neq \Y,\Z$. So $\tau$ usually does nothing to a word, but if the second letter is a $\Y$, it changes it to a $\Z$, and if the second letter is a $\Z$, it changes it to a $\Y$.
Note that since our training set  $D$ contains neither the letters $\Y$ nor $\Z$, then $\tau(D)=D$, as all the words in $D$ satisfy $\tau(w) =w$.

According to Theorem~\ref{thm:main}, to show that $L(D,\Y\Y)=L(D,\Y\Z)$, and therefore that the learning algorithm is not able to generalize the identity effect correctly outside the training set, we just need to show that
\[
L(\tau(D),\tau(w))= L(D,w),
\]
for our $D$ and $w=\Y\Y$. In fact, Theorems~\ref{thm:withRegularization}
 shows that this identity is true for all $D$ and $w$ for certain algorithms and encodings of the inputs. 
A key point is how words are encoded, which then determines the structure of the matrix $\mathcal{T}$, and therefore which results from the previous section are applicable.
We will obtain different results for the invariance of a learning algorithm depending on the properties of $\mathcal{T}$.

First, suppose that letters are encoded using \emph{one-hot} encoding; in this case each letter is represented by a 26-bit vector with a 1 in the space for the corresponding letter and zeros elsewhere. Letting $\mathbf{e}_i$ be the $i$th standard basis vector then gives that $\A$ is encoded by $\mathbf{e}_1$, $\B$ encoded by $\mathbf{e}_2$, etc.  Each input word is then encoded by a 52-bit vector consisting of the two corresponding standard basis vectors concatenated. With this encoding the transformation $\tau$ then just switches the last two entries of the input vector, and so the transformation matrix $\mathcal{T}$ is a permutation matrix. This gives the strongest possible results in our theory:  we can apply Theorem~\ref{thm:withRegularization} with either $\ell_1$ or $\ell_2$ regularization and obtain invariance of the algorithm.
Likewise, Theorem~\ref{thm:SGD} shows that classifiers trained with stochastic gradient descent and $\ell_2$ regularization are also invariant to $\tau$. 
The transformation $\tau$ also satisfies the assumptions of Theorem~\ref{thm:RNNs}. In fact, $\tau = \mathrm{Id} \otimes \tau_2$, where $\tau_2$ switches the letters $\Y$ and $\Z$, and the data set $D$ is invariant to $\tau_2 \otimes \tau_2$ since $\Y$ and $\Z$ do not appear in $D$. Hence, classifiers based on RNN architectures and trained with SGD (without any regularization on the input weights) are invariant to $\tau$.
These results in turn allow us to use Theorem~\ref{thm:main} to show that such learning algorithms are unable to distinguish between the inputs $\Y\Y$ and $\Y\Z$, and therefore cannot learn identity effects from the data given. \new{In the next section we will numerically investigate whether similar conclusions remain valid for some learners that do not satisfy the assumptions of our theory}. 

Second, suppose instead that letters are encoded as orthonormal vectors of length 26, with the $i$th letter encoded as $\mathbf{x}_i$.
Then in this case the transformation $\tau$ switches the last two coefficients of the second letter vector when expanded in this orthonormal basis. So $\tau$ is an orthogonal transformation (in fact a reflection) and $\mathcal{T}$ is an orthogonal matrix, though not a permutation matrix in general. Theorem~\ref{thm:withRegularization} then implies that we have invariance of the learner with the $\ell_2$ regularization, but not with $\ell_1$ regularization. Theorem~\ref{thm:SGD} shows that we have invariance of the learner with SGD with $\ell_2$ regularization (or no regularization at all, if we set the parameter $\lambda=0$).
Moreover, Theorem~\ref{thm:RNNs} shows that invariance also holds for RNNs trained via SGD and without regularization on the input weights. In fact, the transformation $\tau_2$ switches the last two encoding vectors and leaves all the others unchanged. Therefore, thanks to the orthogonality of the encoding vectors, $\tau_2$ is represented by a symmetric and orthogonal matrix. These results will be confirmed when we use an orthogonal Haar basis encoding of letters in the next section. 

Finally, suppose that letters are encoded using arbitrary linearly independent vectors in $\mathbb{R}^{26}$.  Then we have no results available with regularization, though Theorem~\ref{thm:noRegularization} shows we have invariance of the learner if we don't use regularization and we are able to obtain the unique global minimum of the loss function. However, we now show that we can create adversarial examples if we are allowed to use inputs that consist of concatenation of vectors that do not correspond to letters.

\subsection{Adversarial examples for general encodings}

An adversarial example is an input concocted in order to ``fool" a machine learning system; it is an input that a human respondent would classify one way, but the machine learner classifies in another way that we deem incorrect \citep{dalvi2004adversarial,goodfellow2014explaining,thesing2019ai}). 
One way to view the results of the previous subsection is that we show, in certain circumstances, adversarial example for learners trained to learn the identity effect. Given a training set with no words containing $\Y$ or $\Z$, the learner gives the same rating to $\Y\Y$ and $\Y\Z$, and so at least one of them has an incorrect rating and is therefore an adversarial example.
The example we provided have the appealing feature that the inputs still consist of encodings of two-letter words, but it depends on particular encodings of the letters. However, if we are allowed to input any vectors to the learner, we can find adversarial examples for more general situations.

We suppose that the 26 letters are encoded by  vectors  $\mathbf{x}_i$, $i=1,\ldots,26$ of length $m\geq 26$, and that two-letter words are encoded as vectors of length $2m$ by concatenating these vectors. Let $X=\mathrm{Span}(\{\mathbf{x}_i\}_{i=1}^{24})$. Select two orthogonal vectors $\al, \be$ from the orthogonal complement to $X$ in $\mathbb{R}^m$. Note that $\al$ and $\be$ will likely not encode any letter. 
Let $T$ be any orthogonal transformation  on $\mathbb{R}^m$ that is the identity on $X$ and satisfies $T(\al)=\be$, $T(\be)=\al$. Let $\tau$ be the transformation on words that leaves the first letter unchanged but applies $T$ to the second letter. Since the words in $D$ are encoded by the  concatenation of vectors in $X$, we have $\tau(D)=D$. Since $\tau$ is an orthogonal transformation Theorems~\ref{thm:withRegularization} and \ref{thm:SGD} apply with $\ell_2$ regularization. So the learners described in those theorems satisfy invariance with respect to $\tau$.

This gives us a way to construct adversarial examples, with no special requirements on the encodings of the letters. We define the words $w_1=(\al,\al)$ and $w_2=(\al,\be)$. Since $\tau(w_1)=w_2$, Theorem~\ref{thm:main} tells us that $L(D,w_1)=L(D,w_2)$. So the learner is not able to correctly distinguish whether a word is a concatenation of two strings or not. Arguably, the learner is not able to generalize outside the training set, but it could be objected that such inputs are invalid as examples, since they do not consist of concatenations of encodings of letters.

\section{Numerical Experiments} \label{sec:experiments}

\new{In this section we present numerical experiments aimed at investigating to what extent the conclusions of our theory (and, in particular, of Theorems~\ref{thm:SGD} and \ref{thm:RNNs}) remain valid in more practical machine learning scenarios where some of the assumptions made in our theorems do not necessarily hold.} We consider two different experimental settings corresponding to two different identity effect problems of increasing complexity. In the first experimental setting, we \new{study} the problem of identifying whether a two-letter word is composed by identical letters or not, introduced in the opening paragraph of the paper. In the second setting,  we study the problem of learning whether a pair of grey-scale images represent a two-digit number formed by identical digits or not. In both settings, we consider learning algorithms based on different NN architectures and training algorithms.

After providing the technical specifications of the NN learners employed (Section~\ref{sec:IElearners}), we describe the two experimental settings and present the corresponding results in Sections~\ref{sec:alphabet} (Alphabet) and \ref{sec:MNIST} (Handwritten digits). Our results can be reproduced using the code in the GitHub repository \url{https://github.com/mattjliu/Identity-Effects-Experiments}.

\subsection{Learning algorithms for the identify effect problem}
\label{sec:IElearners}

We consider two types of neural network (NN) learning algorithms for the identity effect problem:
 multilayer feedforward NNs trained using stochastic gradient descent (SGD) and long-short term memory (LSTM) NNs  \citep{hochreiter1997long} trained use the Adam method  \citep{kingma2014adam}.  Both NN learners have been implemented in Keras  \citep{chollet2015keras}. Feedforward NNs were already used in the context of identity effects by \citet{tupper2016learning} and LSTM NNs were considered for learning reduplication effects by \citet{prickett2018seq2seq, prickett2019learning}. In the following, we assume the encoding vectors for the characters (either letters or numbers)  to have dimension $n$. In particular, $n = 26$ for the Alphabet example (Section~\ref{sec:alphabet}) and $n = 10$ for the handwritten digit example (Section~\ref{sec:MNIST}). We describe the two network architectures in detail:

\paragraph{Feedforward NNs} The NN architecture has an input layer with dimension $2n$, i.e.\ twice the length of an encoding vector ($n = 26$ or $n = 10$ in our experiments). We consider models with 1, 2 and 3 hidden layers with 256 units each, as in \citet{tupper2016learning}. A ReLU activation is used for all hidden layers. The final  layer has a single output unit. A sigmoid activation is used in the last layer.
For the training, all weights and biases are randomly initialized according to the random Gaussian distribution $\mathcal{N}(\mu,\,\sigma^2)$ with $\mu = 0$ and $\sigma^2 = 0.0025$.  We train the models by minimizing the binary cross-entropy loss function via backpropagation and SGD 
with a learning rate $l = 0.025$. The batch size is set to 72 (i.e., the  number of training samples per epoch) and the number of training  epochs is 5000. Note that this learning algorithm \new{does not satisfy all} the assumptions of \new{Theorem~\ref{thm:SGD}. In fact, the ReLU activation function makes the loss function non-differentiable and the matrix $\mathcal{T}$ associated with the transformation $\tau$ might not be orthogonal, depending on how we encode letters}. 

\paragraph{LSTM NNs} The LSTM (Long-short Term Memory) architecture considered has the following speficiations. The input layer has shape $(2, n)$ where 2 represents the sequence length and $n$ represents the dimension of an  encoding vector ($n = 26$ or $n = 10$ in our experiments). We consider models with 1, 2 and 3 LSTM layers of 32 units each. We used $\tanh$ activation for the forward step and sigmoid activation for the recurrent step. Dropout is applied to all LSTM layers with a dropout probability of 75\%. The output layer has a single output unit, where sigmoid activation is used.
We  train  the  LSTM models  by  minimizing  the  binary  cross-entropy  loss  function via backpropagation using the Adam optimizer with the following hyperparameters: $\gamma = 0.01$, $\beta_1 = 0.9$ and $\beta_2 = 1$.
The kernel weights matrix, used for the linear transformation of the inputs, as well as all biases, are initialized using the random Gaussian distribution $\mathcal{N}(\mu,\,\sigma^2)$ with $\mu = 0$ and $\sigma^2 = 0.0025$. The recurrent kernel weights matrix, used for the linear transformation of the recurrent state, is initialized to an orthogonal matrix (this is the default in Keras). The batch size is set to 72 (the number of training samples per epoch) the number of training epochs is 1000. \new{Note that this learner does not satisfy all the assumptions of Theorem~\ref{thm:RNNs} since it is trained using Adam as opposed to SGD (a theoretical result for learners trained with Adam is proved in the Appendix).}

\subsection{Experimental setting I: Alphabet}
\label{sec:alphabet}
In the first experiment, we consider the problem of identifying if a two-letter word is composed of two identical letters or not. The same problem has also been studied by 
\citet{tupper2016learning}.
However, here we will consider different  NN architectures and training algorithms (see Section~\ref{sec:IElearners}).

\paragraph{Task and data sets} \label{3.3}
Let the vocabulary $W$ be the set of all two-letter words composed with any possible letters from $\A$ to $\Z$. Let $W_1$ denote the set of all grammatically correct words  (i.e. $\A\A, \B\B, \ldots, \Z\Z$) and let $W_0$ denote the set of all other possible words (which in turn are grammatically incorrect). Given a word $w \in W$, the task is to identify whether it belongs to $W_1$ or not. We assign ratings 1 to words in $W_1$ and 0 to words in $W_0$.
Let $D^{\text{train}}$ denote the training data set, which consists of the 24 labelled words $\A\A, \B\B, \C\C, \ldots \X\X$ from $W_1$ along with 48 uniformly sampled words from $W_0$ without replacement. The learners are first trained on $D^{\text{train}}$ and then tested on the test set $D^{\text{test}}$ consisting of the words $\A\A$, $\x\y$, $\Y\Y$, $\Z\Z$, $\Y\Z$, $\Z\Y$, $\E\Y$ and $\S\Z$, where $\x\y$ is the first word from $D^{\text{train}}$ such that $\x\y \in W_0$ (note that there is nothing special about the choice of the letters $\E$ and $\S$ in the last two test words; they were randomly chosen).

\paragraph{Encodings} We represent each word as the concatenation of the encodings of its two letters, and so the representation of the words is determined by the representation of the letters.
All letter representations used have a fixed length of $n = 26$ (chosen due to the 26 letters that make up our vocabulary $W$). We consider the following three encodings:
\begin{enumerate}
\item \textit{One-hot encoding}. 
This encoding simply assigns a single nonzero bit for each character. Namely, the letters $\A$ to $\Z$ are encoded using the standard basis vectors $\mathbf{e}_1, \ldots, \mathbf{e}_{26} \in \mathbb{R}^{26}$, where $\mathbf{e}_i$ has a 1 in position $i$ and 0's elsewhere.

\item \textit{Haar encoding}. The letters are encoded with the rows of a random $26 \times 26$ matrix sampled from the orthogonal group $O(26)$ via the Haar distribution  (see, e.g., \citet{mezzadri2006generate}). 
With this strategy, the encoding vectors form an orthonormal set. 

\item \textit{Distributed encoding}. Each letter is represented by a random combination of $26$ bits. In a $j$-active bits binary encoding, only $j$ random bits are set to 1 and the remaining $26-j$ bits are equal to 0. In our experiments, we set $j = 3$. Moreover, every combination of bits is ensured to correspond to only one letter. 

\end{enumerate}

In the context of our experiments, all random encodings are randomly re-generated for each trial.
Note that for each encoding the matrix $\mathcal{T}$ associated with the the map $\tau$ \new{defined in \eqref{eq:def_tau_IE}} has different properties. For the one-hot encoding, $\mathcal{T}$ is a permutation matrix (and hence orthogonal) that just switches the last two entries of a vector. For the Haar encoding, $\mathcal{T}$ is an orthogonal matrix. Finally, for the 3-active bit binary, $\mathcal{T}$ does not have any special algebraic properties (recall the discussion in Section~\ref{sec:appl_IE}). \new{In particular, with the one-hot encoding, the transformation $\tau$ defined in \eqref{eq:def_tau_IE} satisfies the assumptions of both Theorems~\ref{thm:SGD} and \ref{thm:RNNs}. With the Haar encoding, $\tau$ satisfies the assumptions of Theorem~\ref{thm:SGD}, but not those of Theorem~\ref{thm:RNNs}, with probability 1. When using the distributed encoding, the transformation $\tau$ in \eqref{eq:def_tau_IE} satisfies neither the assumptions of Theorem~\ref{thm:SGD} nor those of Theorem~\ref{thm:RNNs}.}

\paragraph{Randomization strategy}
We repeat each experiment 40 times for each learner. For each trial, we randomly generate a new training data set $D^{\text{train}}$. In the test set $D^{\text{test}}$, the only random word is $\x\y$, chosen from $D^{\text{train}}$. New encodings are also randomly generated for each trial (with the exception of the one-hot case, which remains constant). The same random seed is set once at the beginning of each learner's experiment (not during the 40 individual experiments). Therefore, the same sequence of 40 random data sets is used for every encoding and every learner.

We now discuss the results obtained using the feedforward and LTSM NN learners described in Section~\ref{sec:IElearners}.

\subsection{Results for feedforward NNs (Alphabet)}

\begin{figure}[ht!]
    \centering
    \includegraphics[width=\textwidth]{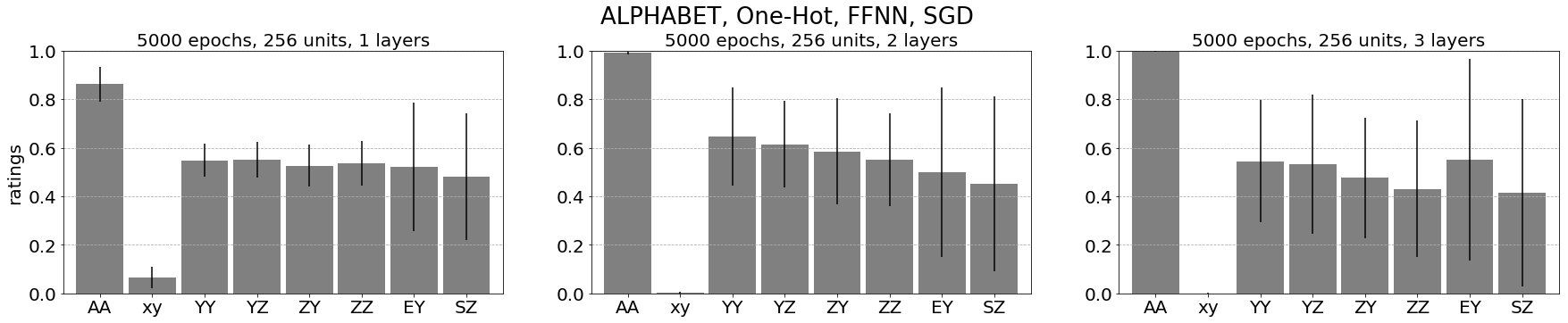}\\
    \includegraphics[width=\textwidth]{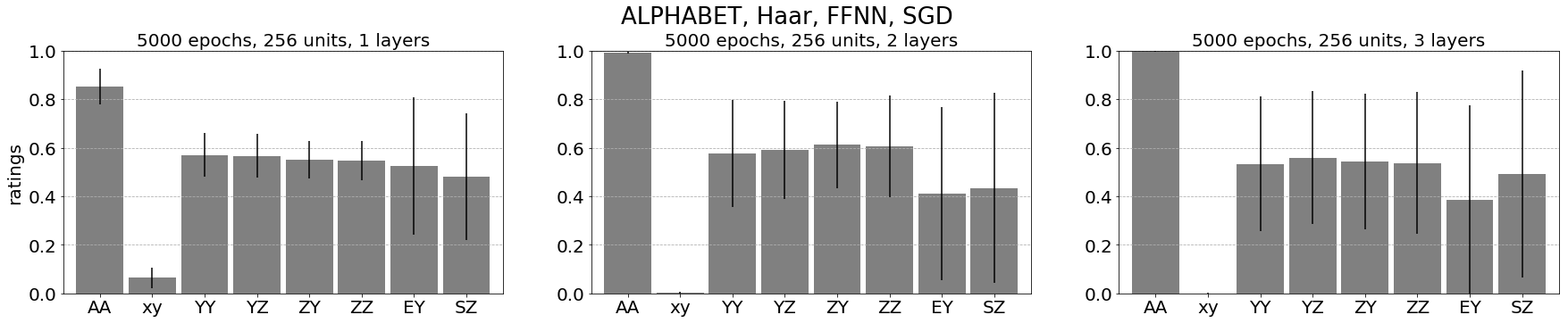}\\
    \includegraphics[width=\textwidth]{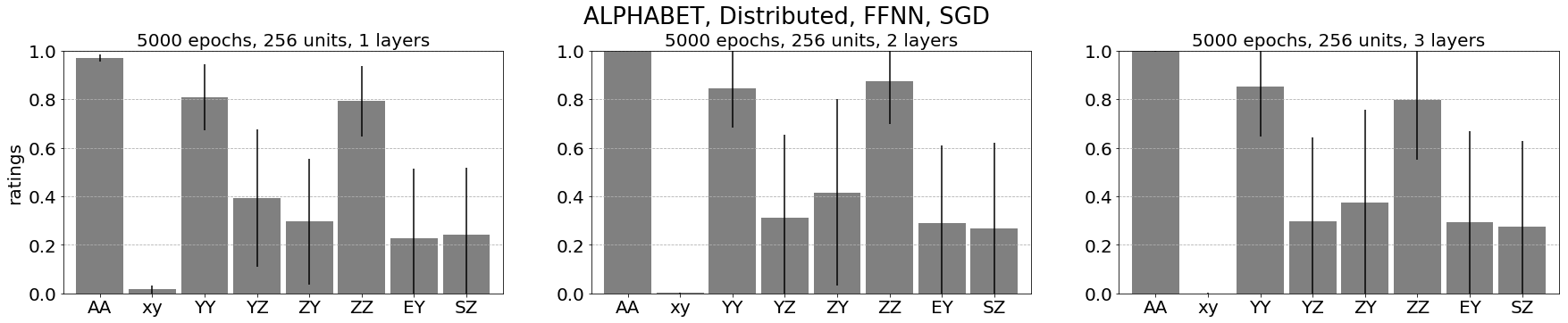}
    \caption{(Alphabet) Ratings produced by SGD-trained feedforward NNs of increasing depth using different encodings. From top to bottom: One-hot encoding, Haar encoding, and distributed encoding. From left to right: 1, 2, and 3 hidden layers. Recall that $\x\y$ denotes the first word from the randomly generated training set $D^{\text{train}}$ such that $\x\y \in W_0$. The first two bars correspond to words in the training set. The last six bars correspond to words not used in the training phase and hence measure the ability of the model to generalize outside the training set.}
    \label{ALPHABET_FFNN_BAR}
\end{figure}

Ratings obtained using SGD-trained feedforward NNs for the Alphabet experiment are shown in Figure~\ref{ALPHABET_FFNN_BAR}. The bars represent the average rating over all 40 trials and the segments represent the corresponding standard deviation.
These results show that feedforward NNs trained via SGD are able to partially generalize to novel inputs only for one of the three encodings considered, namely the distributed encoding (bottom row). We can see this from the fact that these learners assign higher ratings \new{on average} to novel stimuli $\Y\Y$ and $\Z\Z$ than to  novel stimuli $\Y\Z$, $\Z\Y$. The networks trained using the one-hot and Haar encodings (top and middle rows) show no discernible pattern, indicating a complete inability to generalize the identify effects outside the training set. These results follow after all networks are observed to learn the training examples all but perfectly (as evidenced by the high ratings for column $\A\A$ and low ratings for column $\x\y$) with the exception of the 1 layer cases. 

\begin{figure}[ht!]
    \makebox[\textwidth][c]{\includegraphics[width=\textwidth]{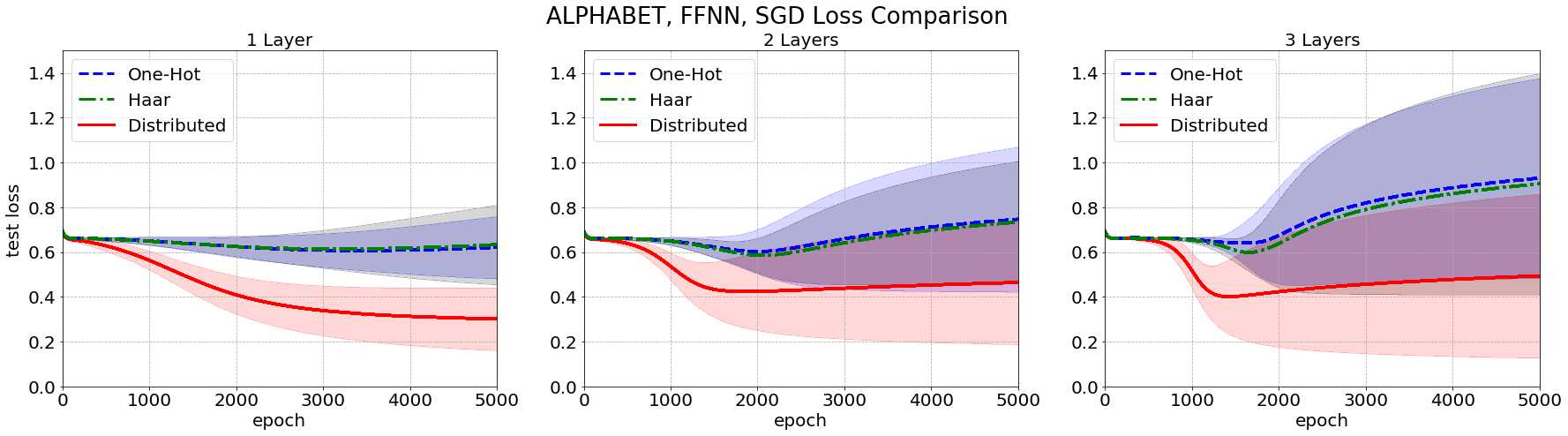}}
    \caption{(Alphabet) Plot of the test loss (binary cross entropy) as a function of the training epoch for feedforward NNs of increasing depth, using  different encodings. From left to right: 1, 2, and 3 hidden layers. Lines and shaded regions represent mean and standard deviation of the test loss across 40 random trials.}
    \label{ALPHABET_FFNN_COMPARISON}
\end{figure}

In Figure~\ref{ALPHABET_FFNN_COMPARISON}, we further compare the three encodings by plotting the test loss as a function of the training epoch. 
Lines represent the mean test loss and shaded areas represent the standard deviation over 40 trials. We see that the mean test loss for the distributed encoding (solid red line) is consistently below the other two lines, corresponding to the one-hot and the Haar encodings (the same pattern also appears with the shaded regions). 

\new{These results seem to suggest that the rating impossibility implied by Theorems~\ref{thm:main} and \ref{thm:SGD} holds for the one-hot and the Haar encodings in the numerical setting considered, despite the fact that the assumptions of Theorem~\ref{thm:SGD} are not satisfied (due to the non-differentiability of the ReLU activation).}

\subsubsection{Results for LSTM NNs (Alphabet)}

\begin{figure}[ht!]
    \centering
    \includegraphics[width=\textwidth]{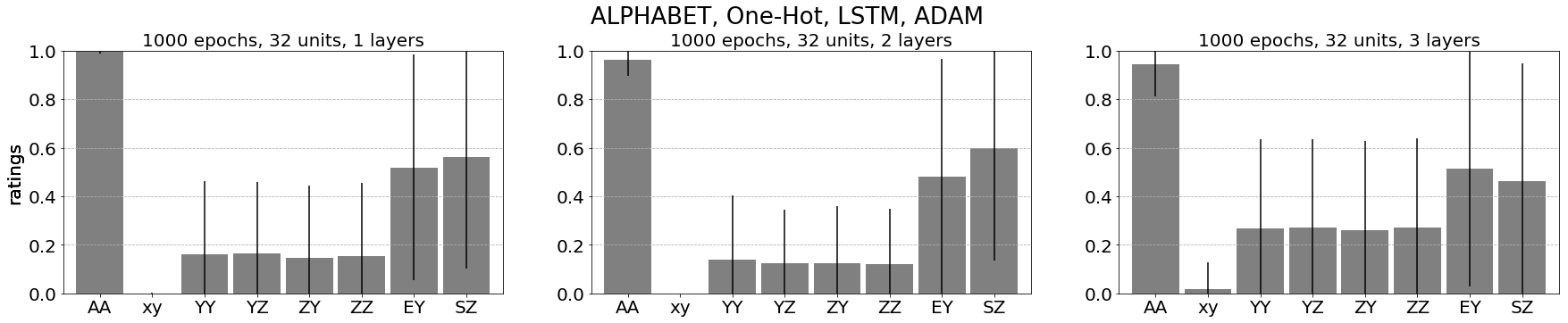}\\
    \includegraphics[width=\textwidth]{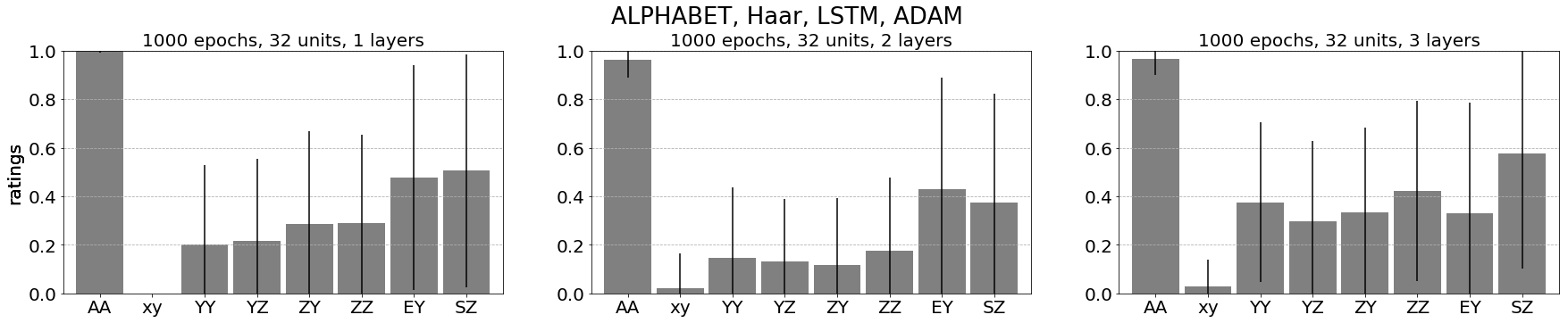}\\
    \includegraphics[width=\textwidth]{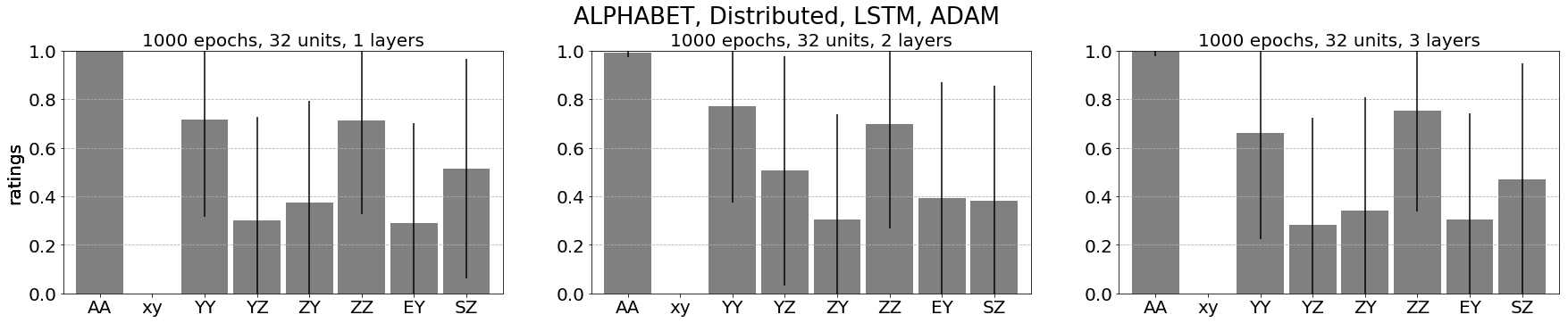}
    \caption{(Alphabet) Same bar plots as in Figure~\ref{ALPHABET_FFNN_BAR} for Adam-trained LSTM NNs.}
    \label{ALPHABET_LSTM_BAR}
\end{figure}

Figure~\ref{ALPHABET_LSTM_BAR} shows ratings produced by Adam-trained LSTM NNs of increasing depth and using different encodings. 
The trend observed is similar to the one in the one obtained using SGD-trained feedforward NNs, with some key differences. In fact, we see a partial ability of these learners to generalize the identity effect outside the training set using the distributed encoding (bottom row) and a complete inability to do so when the one-hot and the Haar encodings are employed (top and middle rows). We note, however, that the pattern suggesting partial ability to generalize in the distributed case is much less pronounced than in the feedforward case. Furthermore, the learning algorithms seems to promote ratings closer to 0 in the one-hot and the Haar cases with respect to the feedforward case, where ratings assigned to words in the test set are closer to 0.5.

\begin{figure}[ht!]
    \makebox[\textwidth][c]{\includegraphics[width=\textwidth]{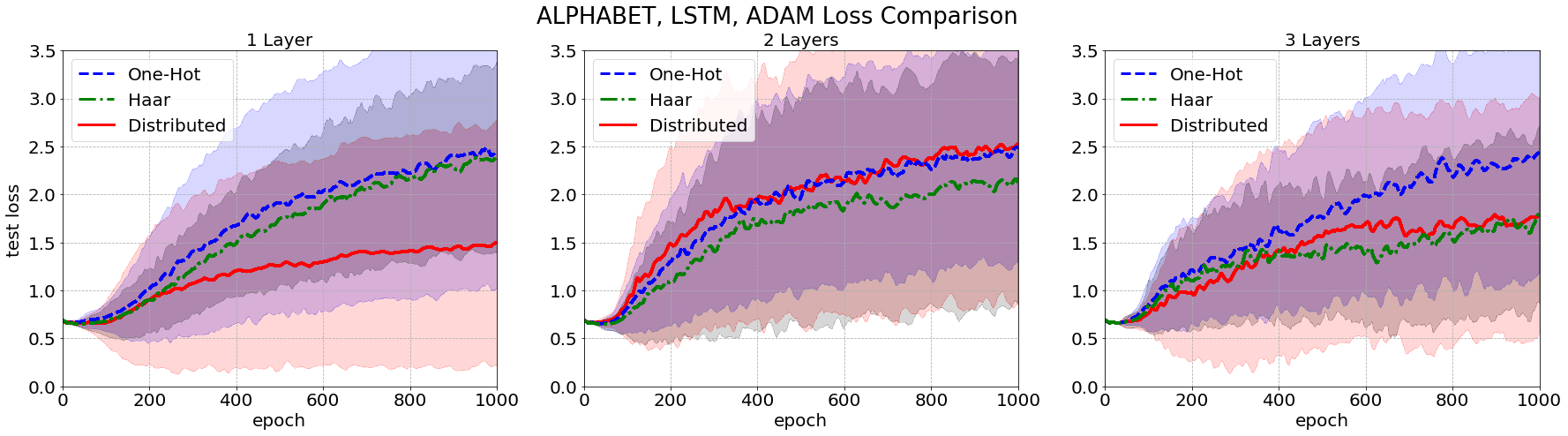}}
    \caption{(Alphabet) Same plot as in Figure~\ref{ALPHABET_FFNN_COMPARISON} for LSTM NNs.}
    \label{ALPHABET_LSTM_COMPARISON}
\end{figure}

Figure~\ref{ALPHABET_LSTM_COMPARISON} shows the mean test loss as a function of the training epoch for different encodings. 
We can now observe that only in the 1 layer case  the distributed mean loss curve (solid red line) lies consistently below the other curves. This seems to suggest that the depth of the LSTM negatively impacts the model's ability to generalize. 

\new{Let us once again comment these results in view of our theory. The rating impossibility implied by our theory (in this case, obtained by combining Theorems~\ref{thm:main} and \ref{thm:RNNs}) seems to hold in the LSTM setting with both the one-hot and Haar encodings. Comparing this setting with the feedforward NN case, there is a wider gap between our theoretical assumptions and the numerical setting. In fact, the assumptions of Theorem~\ref{thm:RNNs} are not satisfied because the learner is trained using Adam as opposed to SGD. In addition, for the Haar encoding, the matrix $\mathcal{T}$ associated with the transformation $\tau$ in \eqref{eq:def_tau_IE} does not fall within the theoretical framework of Theorem~\ref{thm:RNNs}.}

\subsection{Experimental setting II: Handwritten digits}

\label{sec:MNIST}

The identity effect problem considered in the second experimental setting is similar to that of the Alphabet experiment (Section~\ref{sec:alphabet}), but we consider pairs handwritten digits instead of characters. Given two images of handwritten digits, we would like to train a model to identify whether they belong to the same class (i.e., whether they represent the same abstract digit $0, 1, \ldots 9$) or not, in other words, if they are ``identical'' or not. Therefore, being an identical pair is equivalent to identifying if a 2-digit number is palindromic. \new{Considerations analogous to those made in Section~\ref{sec:appl_IE} are valid also in this case, up to replacing the definition of the transformation $\tau$ defined in \eqref{eq:def_tau_IE} with
\begin{equation}
\label{eq:def_tau_MNIST}
\tau(x8)=x9, \ \ \ \tau(x9)=x8, \ \ \ \tau(xy)=xy, 
\end{equation} 
for all digits $x$ and $y$ with $y \neq 8,9$. However, a crucial difference with respect to the Alphabet case is that the encoding used to represent digits is itself the result of a learning process.} Images of handwritten digits are taken from the popular MNIST data set of \citet{lecun2010mnist}. 

\subsubsection{Learning algorithm: Computer vision and identity effect models}

\label{sec:MNIST_learners}

We propose to solve the problem by concatenating and combining two distinct models: one for the image classification task, which entails the use of a computer vision model and another for the identity effects part, whose purpose is to identify if two digits belong to the same class or not. 

The \textit{Computer Vision} (\textit{CV}) model takes as input a given $28 \times 28$ grey scale image from the MNIST data set. The output is a $10$-dimensional vector (for each of the $10$ MNIST classes) produced by a final softmax prediction layer. As such, the main purpose of the CV models is to encode an MNIST image into a $10$-dimensional probability vector. This \new{learned} encoding can be thought of as the one-hot encoding corrupted by additive noise. \new{Due to the learned nature of the CV encoding, the matrix $\mathcal{T}$ associated with the transformation $\tau$ in \eqref{eq:def_tau_MNIST} is not orthogonal nor a permutation matrix. Therefore, the assumptions involving $\tau$ in Theorems~\ref{thm:SGD} or \ref{thm:RNNs} are not satisfied.}

\begin{figure}[ht!]
    \makebox[\textwidth][c]{\includegraphics[width=\textwidth]{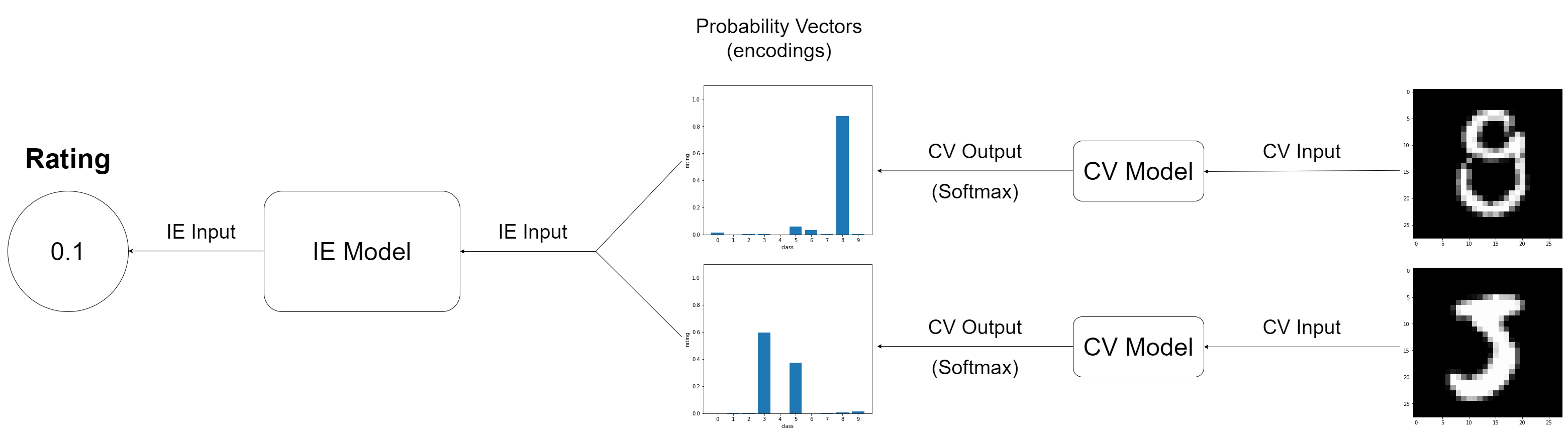}}
    \caption{(Handwritten digits) Diagram of the learning model used in the handwritten digits experiment. The model is obtained by concatenating a computer vision (CV) model and an identity effect (IE) model. From right to left: The model takes a pair of images as input (right); then, the CV model classifies them or, equivalently, encodes them as two 10-dimensional probability vectors (center); finally, the IE model assigns a rating from 0 to 1 to the pair of probability (or encoding) vectors to identify whether the images represent identical digits or not (left).}
    \label{MNIST_model_diagram}
\end{figure}

The \textit{Identify Effect} (\textit{IE}) model takes a 20-dimensional vector (i.e., the concatenation of two 10-dimensional vectors output by the CV model) and returns a single value (the rating) predicting whether or not the pair is identical. Figure~\ref{MNIST_model_diagram} illustrates the how the CV and IE models are combined in the handwritten digits setting. One of the main objectives of this experiments is to understand the interplay between the training of the CV and the IE model. 

We now describe the architectures and the training algorithms considered for the CV and the IE models.
\paragraph{CV model specifications}
We use the official Keras ``Simple MNIST convnet'' model \citep{simpleMNISTconvnet}, formed by the following components: (i) A 2D convolutional layer with $32$ filters (output dimension of $32$). The kernel size is $3\times 3$ with a stride of $1 \times 1$. This is applied on an input of $28 \times 28 \times 1$, which gives an output of $26 \times 26 \times 32$. ReLU activation is used. (ii) A 2D convolutional layer with $64$ filters. The kernel size is $3\times 3$ with a stride of $1\times 1$. This gives an output of $24\times 24\times 64$. ReLU activation is used. (iii) A 2D max pooling layer (max filter) with a pool size of $2\times 2$ (halving on both axis). Output size of $12\times 12 \times 64$. Dropout is applied to this layer with a probability of $0.25$. (iv) The previous output is flattened into a single $9216$ dimension layer and feed into a $128$ unit layer. ReLU activation is used and dropout is applied to this layer with a probability of $0.5$. (v) A final $10$-dimensional softmax output layer. We train the CV model by minimizing the categorical cross-entropy loss function via backpropogation and the Adadelta optimizer \citep{zeiler2012adadelta} with $lr=0.001$ and $\rho=0.95$. Kernel weights are initialized using the uniform initializer by \citet{Glorot2010UnderstandingTD}. Biases are initilized to 0. The batch size is set to $128$.

\paragraph{IE model specifications}

The IE models are feedforward and LSTM NNs like those described in Section~\ref{sec:IElearners}, with $n = 10$ (encoding vectors have length $10$). Moreover, we use the Adam optimizer instead of SGD to train the feedforward NNs with the following hyperparameters: $\gamma = 0.01$, $\beta_1 = 0.9$ and $\beta_2 = 1$. The batch size is also changed to $2400$ (the size of the training set). This modification was made to speed up simulations thanks to the faster convergence of Adam with respect to SGD. Using SGD leads to similar results.

\subsubsection{Construction of the training and test sets}

The standard MNIST data set contains a training set of 60,000 labelled examples and a test set of 10,000 labelled examples. Let us denote them as
$$
D^{\text{train}}_{\text{MNIST}} = \{(\X_i, d_i)\}_{i = 1}^{60000}, \quad
D^{\text{test}}_{\text{MNIST}} = \{(\Y_i, e_i)\}_{i = 1}^{10000},
$$
where, for every $i$, $\X_i,\Y_i\in \mathbb{R}^{28 \times 28}$ are grey-scale images of handwritten digits, with labels $d_i,e_i \in \{0,\ldots,9\}$, respectively. The CV model is trained on the MNIST training set $D^{\text{train}}_{\text{MNIST}}$. Given a trained CV model, we consider the corresponding \textit{CV model encoding} 
\begin{equation}
\label{eq:def_CV_encoding}
\mathcal{E}_{\text{CV}} : \mathbb{R}^{28 \times 28} \to [0,1]^{10}.
\end{equation}
For any image $\X \in \mathbb{R}^{28 \times 28}$, the map $\mathcal{E}_{\text{CV}}$ returns a $10$-dimensional  probability vector $p = \mathcal{E}_{\text{CV}}(\X) \in [0,1]^{10}$ obtained by applying the softmax function to the output generated by the CV model from the input $\X$ (recall Section~\ref{sec:MNIST_learners} about the CV model architecture see and Figure~\ref{MNIST_model_diagram} for a visual intuition). 

For the IE model, we define the training and test sets as follows: 
$$
D_{\text{IE}}^{\text{train}} = \{(\mathcal{E}_{\text{CV}}(\widetilde{\X}_i^1), \mathcal{E}_{\text{CV}}(\widetilde{\X}_i^2)), r_i)\}_{i = 1}^{2400},\quad
D_{\text{IE}}^{\text{test}} = \{(\mathcal{E}_{\text{CV}}(\widetilde{\Y}_i^1), \mathcal{E}_{\text{CV}}(\widetilde{\Y}_i^2)), s_i)\}_{i = 1}^{10},
$$
where the images $\widetilde{\X}_i^k, \widetilde{\Y}_i^k \in \mathbb{R}^{28 \times 28}$ are randomly sampled from the MNIST test set $D^{\text{test}}_{\text{MNIST}}$ according to a procedure described below. The rating $r_i$ is equal to  $1$ if the images $\widetilde{\X}_i^1$ and $\widetilde{\X}_i^2$ correspond to identical digits (according to the initial labelling in the MNIST test set $D^{\text{test}}_{\text{MNIST}}$) and $0$ otherwise. The ratings $s_i$ are defined accordingly. The rationale behind the number of training examples ($=2400$) and test examples ($=10$) will be explained in a moment. Since the feedforward IE model must evaluate two digits at a time, the two corresponding probability vectors are concatenated to form a $20$-dimensional input. In the LSTM case, the two $10$-dimensional vectors are fed in as a sequence to the IE model.

Let us provide further details on the construction of $D_{\text{IE}}^{\text{train}}$. Let $W$ be the set of all two-digit numbers formed by digits from 0 to 9. We define the set $W_1$ as the set of all two-digit numbers formed by identical digits (i.e. $00, 11, \ldots, 99$) and $W_0$ as the set of all other possible two-digit numbers. Then, $D_{\text{IE}}^{\text{train}}$ is constructed in two steps: 
\begin{enumerate}
    \item[Step 1.] For every digit $n = 0, \ldots, 7$, we sample $10$ images labelled as $n$ uniformly at random from the MNIST test set $D^{\text{test}}_{\text{MNIST}}$. This leads to $80$ random images in total. We call the set formed by these images $D^{\text{test}}_{\text{MNIST},\leq 7}$. The pairs forming the set $D_{\text{IE}}^{\text{train}}$ are composed by CV model encodings of random pairs of images in $D^{\text{test}}_{\text{MNIST},\leq 7}$. 
    \item[Step 2.] In order to keep the same ratio between the number of training pairs in $W_0$ and those in $W_1$ as in the Alphabet experiment (i.e., a $1:2$ ratio), we use of all possible identical pairs and only keep $2/7$ of all possible nonidentical pairs from $D^{\text{test}}_{\text{MNIST},\leq 7}$. This yields $8\cdot10^2 =800$ identical pairs (belonging to $W_1$) and $8\cdot7\cdot10^2 \cdot 2/7 = 1600$ nonidentical pairs (belonging to $W_0$), for a total of $2400$  pairs of images. The training examples in $D^{\text{train}}_{IE}$ are the CV model encodings of these $2400$ image pairs.
\end{enumerate}

Let us now define the test set $D^{\text{test}}_{IE}$. First, we choose random images $\X$, $\Y$, $\X'$, $\Y'$, $\eight$, and $\nine$ from $D^{\text{test}}_{\text{MNIST}}$ as follows:
\begin{itemize}
    \item [$\X$, $\Y$:] Two images of distinct digits from $0$ to $7$ sampled uniformly at random from the set $D_{\text{MNIST}, \leq 7}^{\text{test}}$ defined in Step 1 above; 
    \item [$\X'$, $\Y'$:] Two images of distinct digits from $0$ to $7$ sampled uniformly at random from $D_{\text{MNIST}}^{\text{test}}$ that do not belong to $D_{\text{MNIST}, \leq 7}^{\text{test}}$; 
    \item [$\eight$, $\nine$:] Two random images labelled as $8$ and $9$ from $D^{\text{test}}_{\text{MNIST}}$ (hence, not used in $D_{\text{IE}}^{\text{train}}$ by construction).
\end{itemize}
The images $\X$, $\Y$, $\X'$, $\Y'$, $\eight$, and $\nine$ are then used to construct ten pairs $(\X,\X)$, $(\X, \Y)$, $(\X', \X')$, $(\X',\Y')$, $(\eight,\eight)$, $(\eight,\nine)$, $(\nine,\eight)$, $(\nine,\nine)$, $(\nine,\nine)$, $(\X',\eight)$, $(\X',\nine)$. The CV model encoding of these pairs form the test set  $D^{\text{test}}_{IE}$. In order to simplify the notation, we will omit the brackets and the map $\mathcal{E}_{\text{CV}}$ when referring to the elements of $D^{\text{test}}_{IE}$. For example, the pair $(\mathcal{E}_{\text{CV}}(\X'), \mathcal{E}_{\text{CV}}(\eight))$ will be denoted as  $\X'\eight$. Therefore, we have 
$$
D^{\text{test}}_{IE}
= \{\X\X, \X\Y, \X'\X', \X'\Y', \eight\eight, \eight\nine, \nine\eight, \nine\nine, \X'\eight, \X'\nine\}.
$$
The first two test pairs $\X\X, \X\Y$ are used to measure the performance of the IE model inside the training set. The role of the pairs $\X'\X', \X'\Y'$ is to assess the ability of the IE model to generalize to new images of previously seen digits (from $0$ to $7$). Finally, the pairs $\eight\eight, \eight\nine, \nine\eight, \nine\nine, \X'\eight, \X'\nine$ are used gauge to what extend the IE model can fully generalize outside the training set (both in terms of unseen images and unseen digits).

\subsubsection{Training strategies and corresponding encodings}
\begin{figure}[ht!]
\centering
    \includegraphics[width=0.8\textwidth]{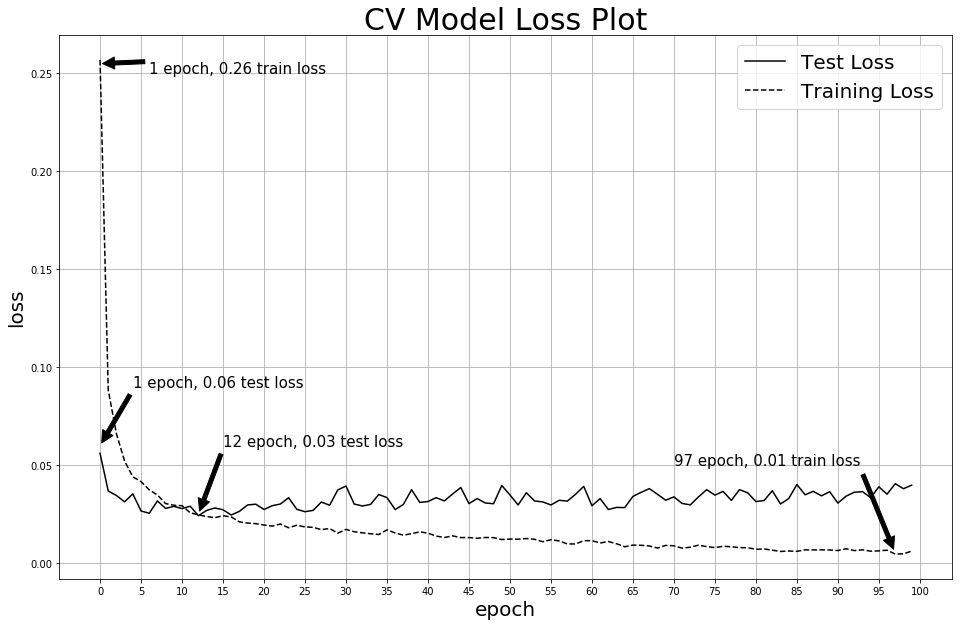}
    \caption{(Handwritten digits) Training and test loss (categorical cross entropy) as a function of the training epoch for the CV model. Arrows indicate the global extrema of the training and test loss over 100 epochs. Different stopping criteria lead to different CV model encodings $\mathcal{E}_{\text{CV}}$ defined in \eqref{eq:def_CV_encoding}. In turn, this corresponds to perturbed versions of the one-hot encoding by different amounts of additive noise.}
    \label{CV_loss}
\end{figure}

By construction, the CV model encoding $\mathcal{E}_{\text{CV}}$ defined in \eqref{eq:def_CV_encoding} depends how the CV model is trained. Moreover, the same is true for the sets $D_{\text{IE}}^{\text{train}}$ and $D_{\text{IE}}^{\text{test}}$ used to train and test the IE model, respectively. Here, we consider two possible scenarios: the undertrained and the optimally-trained case. In the undertrained case, we only train the CV model for 1 epoch. In the optimally-trained case, we train the CV model for 12 epochs, corresponding to the minimum test loss over the 100 epochs considered in our experiment. This is illustrated in Figure~\ref{CV_loss}.  
Recalling that $\mathcal{E}_{\text{CV}}$ can be thought of as a perturbation of the one-hot encoding by additive noise, the undertrained scenario corresponds to perturbing the one-hot encoding by a large amount of additive noise. In the optimally-trained scenario, the CV model encoding is closer to the true one-hot encoding.

\subsubsection{Results for feedforward NNs (handwritten digits)}

\begin{figure}[ht!]
    \centering
    \includegraphics[width=\textwidth]{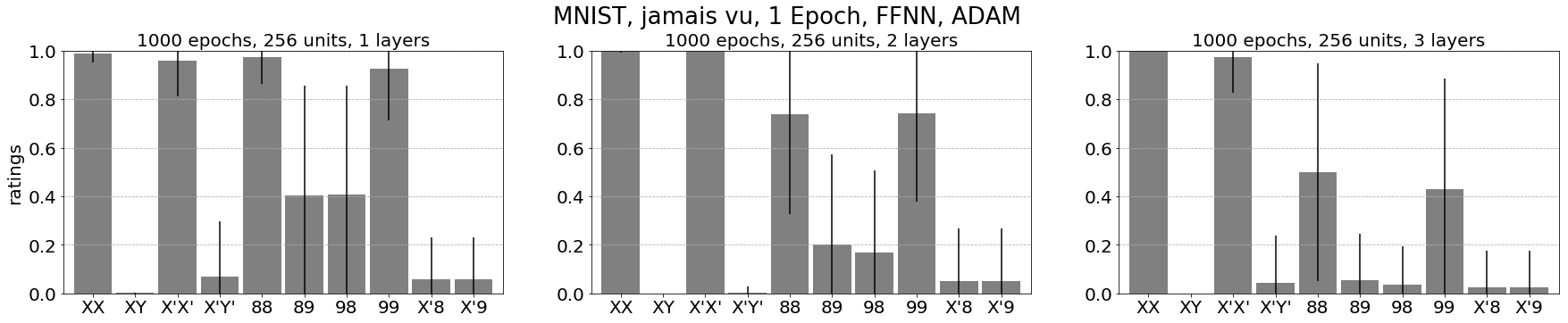}\\
    \includegraphics[width=\textwidth]{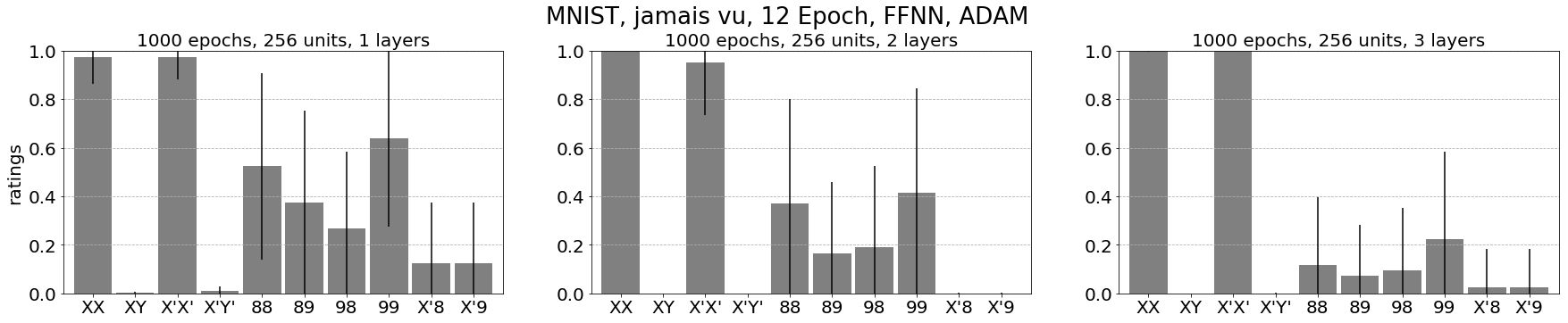}
    \caption{(Handwritten digits) Ratings produced by IE models with feedforward NN architectures of increasing depth for  different training levels of the CV model. From top to bottom:  Undertrained CV model (1 epoch) and optimally-trained CV model (12 epochs), chosen as in Figure~\ref{CV_loss}. The first two bars correspond to image pairs $\X\X$ and $\X\Y$ used during the IE model training. The third and the fourth bars represent image pairs $\X'\X'$ and $\X'\Y'$ not used to train the IE model, but corresponding to digits from $0$ to $7$ that the IE model was trained on (with different images). The last six bars correspond to numbers and images not used in the IE model training (i.e., where at least one of the digits is an 8 or a 9). Therefore, the first two bars measure the performance of the IE model on the training set; the third and the fourth bars represent the ability of the IE model to generalize to unseen images (but already seen digits); the last six bars  measure the ability of the model to generalize outside the training set (in terms of both images and digits).}
    \label{ALPHA0_MNIST_FFNN_BAR}
\end{figure}

The results for feedforward NNs with undertrained and optimally-trained CV models are shown in Figure~\ref{ALPHA0_MNIST_FFNN_BAR}. 
Similarly to the the Alphabet experiment, the bar plots correspond to average ratings computed over 40 random trials. The bar plots show that the shallow (1 layer) undertrained CV model learner (top left plot) performs the best (as evidenced by the high ratings for the pairs $\eight\eight$ and $\nine\nine$). We can also observe that using an undertrained CV model (top row) consistently leads to a better ability to generalize outside the training set for the IE model, if compared with the case of an  optimally-trained CV model (bottom row). This is especially evident in the 3 layer case (right-most column), where there is only a weakly discernible pattern in the model outputs for the optimally-trained CV model. This observation is aligned with our theoretical results. In fact, in the optimally-trained scenario, the CV model encoding is closer to the one-hot econding (which, in turn, makes the task of learning an identity effect impossible, due to its orthogonality, \new{in view of Theorem~\ref{thm:SGD}}). The partial generalization effect is due to the fact that the CV model is a perturbation of the one-hot encoding and the additional noise is what makes it possible for the IE model to break the ``orthogonality barrier''. We also note that the IE model is able to perform extremely well on previously seen digits (from the scores in the first 4 bars of each plot), even if the corresponding images were not used in the training phase.

\begin{figure}[ht!]
    \makebox[\textwidth][c]{\includegraphics[width=\textwidth]{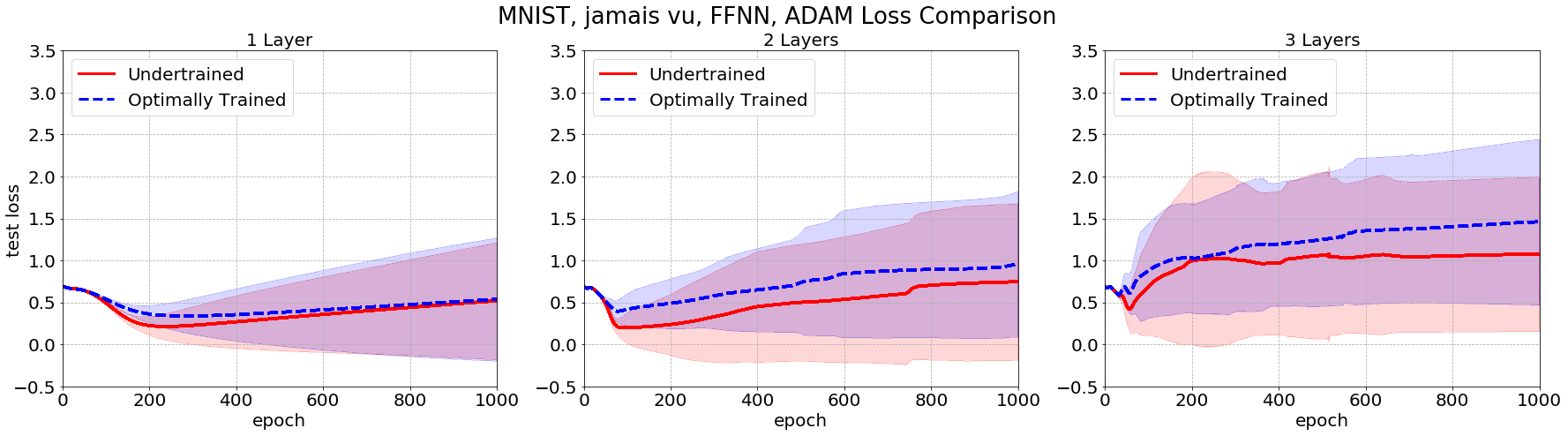}}
    \caption{(Handwritten digits) Plot of the test loss (categorical cross entropy) as a function of the training epoch for IE models based on feedforward NN architectures of increasing depth, for different training levels of the CV model (i.e., undertrained and optimally trained). From left to right: 1, 2, and 3 hidden layers. Lines and shaded regions represent mean and standard deviation of the test loss across 40 random trials. }
    \label{ALPHA0_MNIST_FFNN_COMPARISON}
\end{figure}

These observations are by confirmed Figure~\ref{ALPHA0_MNIST_FFNN_COMPARISON}, showing the evolution of the test error associated with the IE model as a function of the training epoch.
Indeed, the solid curves, representing the undertrained CV model, are consistently  below the dashed curves, representing the optimally-trained CV model.

\subsubsection{Results for LSTM NNs (handwritten digits)}

\begin{figure}[ht!]
    \centering
    \includegraphics[width=\textwidth]{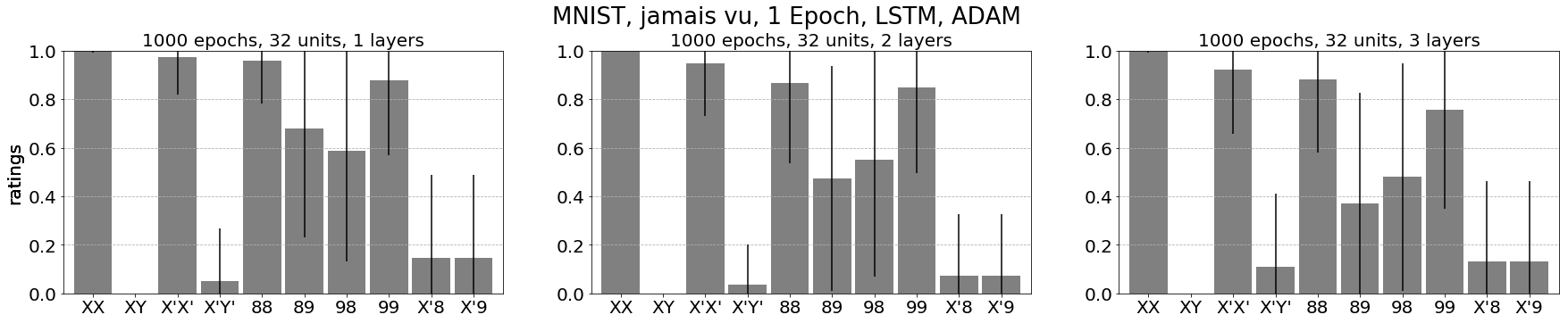}\\
    \includegraphics[width=\textwidth]{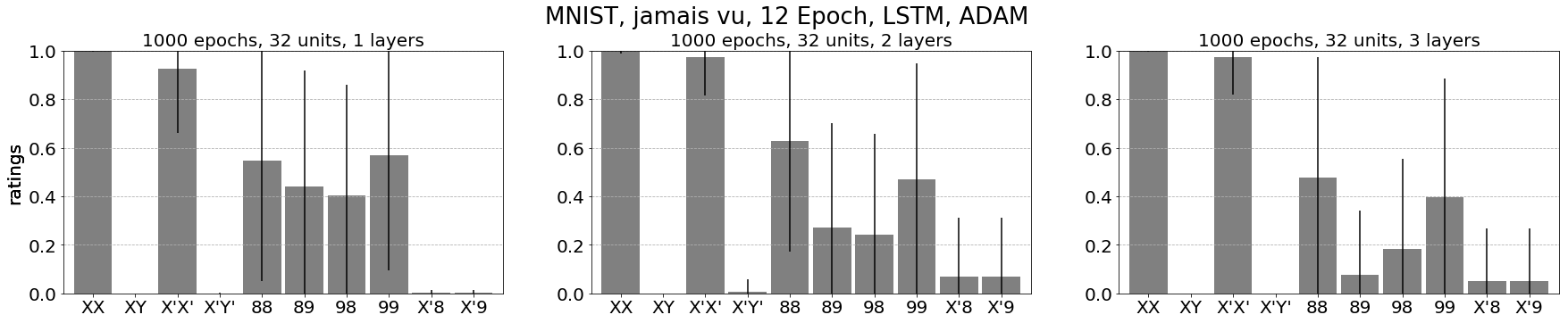}

    \caption{(Handwritten digits) Same bar plots as in Figure~\ref{ALPHA0_MNIST_FFNN_BAR} for IE models based on LSTM NN architectures.}
    \label{ALPHA0_MNIST_LSTM_BAR}
\end{figure}

The results for IE models based on LSTM NN architecrues  are shown in Figure~\ref{ALPHA0_MNIST_LSTM_BAR}. 
From these results, see that the performance of the LSTM models is similar to the feedforward models. It is worth observing that the undertrained CV model (top row) produces high average scores for $\eight\eight$ and $\nine\nine$ in the test set. However, the average scores for all other numbers are also higher. The same holds in the optimally-trained case (bottom row).

\begin{figure}[ht!]
    \makebox[\textwidth][c]{\includegraphics[width=\textwidth]{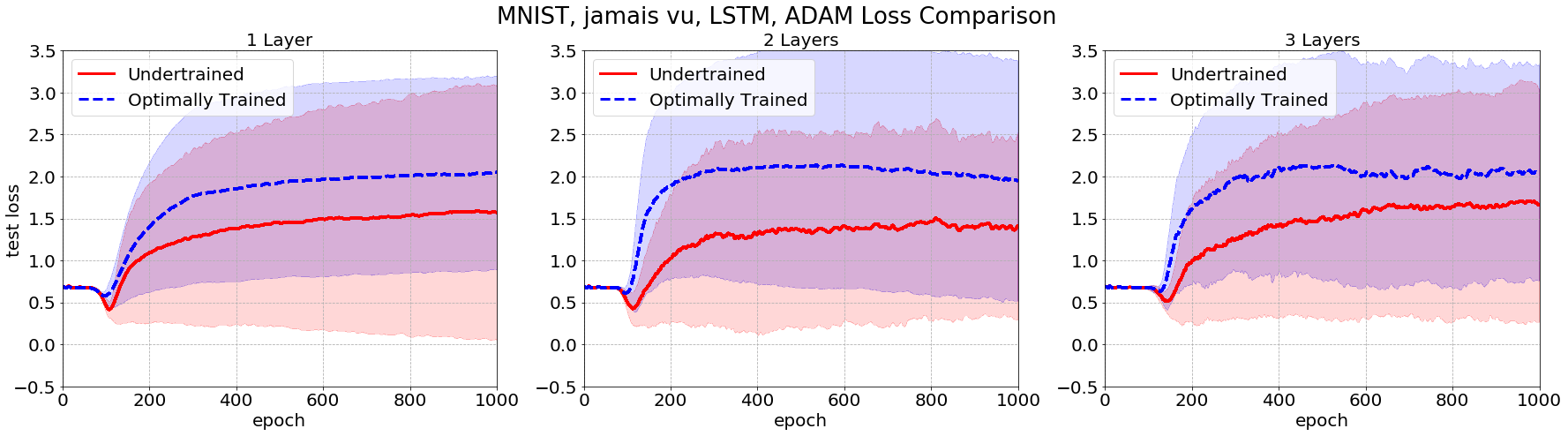}}
    \caption{(Handwritten digits) Same plots as in Figure~\ref{ALPHA0_MNIST_FFNN_COMPARISON} for IE models based on LSTM NN architectures.}
    \label{ALPHA0_MNIST_LSTM_COMPARISON}
\end{figure}

Figure~\ref{ALPHA0_MNIST_LSTM_COMPARISON} shows the evolution of the test loss as a function of the training epoch for the IE models. 
We see again that the solid lines representing  the mean test losses for undertrained CV models are consistently below the dashed lines, representing the mean test losses for optimally-trained CV models.

\new{These numerical results parallel the conclusions of our theory. In fact, in the optimally-trained scenario, the CV model encoding gets closer to the one-hot encoding (for which the transformation $\tau$ in \eqref{eq:def_tau_MNIST} satisfies the assumptions of Theorem~\ref{thm:RNNs}), and our numerical results show an increased difficulty for the IE model to generalize outside the training set. This observation is consistent with the rating impossibility implied by Theorems~\ref{thm:main} and \ref{thm:RNNs} (with the proviso that the IE learner does not formally satisfy the assumptions of Theorem~\ref{thm:RNNs} due to the use of Adam for training -- see also the Appendix).}

\subsubsection{The ``jamais vu" and the ``déjà vu'' scenarios} 
\label{sec:jamais_vu_deja_vu}

We conclude by noting that the definitions of $D_{\text{IE}}^{\text{train}}$ and $D_{\text{IE}}^{\text{test}}$ considered here  correspond to a ``jamais vu'' (i.e., ``never seen'') scenario, where the IE model is trained and tested only on examples that the CV model was not trained on. It is also possible to consider a ``d\'ej\`a vu'' (i.e., ``already seen'') scenario, where the IE model is trained with digits from the MNIST training set $D^{\text{train}}_{\text{MNIST}}$, already used to train the CV model. In this paper, we only show results for the ``jamais vu'' setting, although we run similar experiments in the ``d\'ej\`a vu'' case. In the ``d\'ej\`a vu" case, the CV model is undertrained at 1 epoch (corresponding to the largest training error in Figure~\ref{CV_loss}) and optimally trained at 97 epochs (corresponding to the minimum training error in Figure~\ref{CV_loss}). It is possible to see that in the ``d\'ej\`a vu" scenario, it is even more difficult for the IE model to learn the identity effect, especially in the optimally-trained case since the CV model encoding is very close to the one-hot encoding. For further details, we refer to our GitHub repository \url{https://github.com/mattjliu/Identity-Effects-Experiments}.

\section{Conclusion}

Let us go back to the identity effect problem introduced in the opening paragraph. We see agreement between our theoretical predications, discussed in Section~\ref{sec:identity}, and the numerical experiments of Section~\ref{sec:alphabet} (Alphabet setting). Our theory predicted that when the encoded letters for different vectors are orthogonal (as they are with one-hot and Haar encodings), then since the transformation $\tau$  is an orthogonal transformation, the learner will not be able to distinguish between the inputs $\Y\Y$ and $\Y\Z$. In accordance with predictions, we numerically observed a complete inability of feedforward and LSTM NNs to generalize this type of identity effects outside the training set with these orthogonal encodings regardless of their depth (from 1 to 3) and of the training algorithm employed (SGD or Adam).

Our theory has nothing to say about the case of the 3-bit active encoding, because in that case $\tau$ is not orthogonal, and our theorems do not apply. However, in this case we  showed the existence of adversarial examples able to ``fool'' the learning algorithm using encodings that are orthogonal vectors corresponding to letters from $\A$ to $\X$. In this case, our numerical experiments showed that even though the network is not able to give the correct answer of $1$ for $\Y\Y$ and $0$ for $\Y\Z$, and so not be said to learn the generalization perfectly, it does give a higher rating on average to  $\Y\Y$ than to $\Y\Z$.
We leave it to the reader to decide if this constitutes an exception to the claim that learners need to instantiate variables in order to generalize algebraic rules outside the training set, supported by \citet{marcus1999}.

Our results hew closely to those of \cite{prickett2019learning}; see also \cite{prickett2018seq2seq}. There the authors train a variable-free neural network to perform reduplication, the process where a linguistic element is repeated from the input to the output. Following the experimental work of \cite{marcus1999}, they trained the network on many examples of the pattern ABB, where A and B are substituted with syllables. The network is then tested by seeing if it can predict that the third syllable of a string such as ``li na $\rule{0.35cm}{0.15mm}$'' should be ``na'', even when not exposed to this input before. The authors found that their network could perform partial generalization when the novel inputs included new syllables or new segments, but could not generalize to new feature values. The reason for this is that feature values were encoded in their model via a localist representation, and introducing a new feature value was like expecting the network to learn a function depending on a bit that was always set to zero in the training data, just like the localist representation in our set-up. Since novel segments were composed of multiple novel feature values, this corresponds to our 3-bit active encoding, where apparently learning can be extended imperfectly to new combinations of already seen segments.

Our results and those of \cite{prickett2019learning} continue a theme that is well known in connectionist literature: when representations of novel inputs overlap with representations in training data, networks are able to generalize training to novel inputs. See \cite{mcclelland1999} for a discussion of this point in the context of identity effects.

Furthermore, in the handwritten digits experiment (Section~\ref{sec:MNIST}), we considered the problem of learning whether a pair of images represents identical digits or not. This setting required the introduction of more complex learning algorithms, obtained by concatenating a Computer Vision (CV) and an Identity Effect (IE) model (see Figure~\ref{MNIST_model_diagram}). In this case, the encoding is given by the probability vectors generated as softmax outputs of the CV model and can be though of as a one-hot encoding plus additive noise. In accord with our theory, we observed that generalizing the identity effect outside the training set becomes more difficult as the encoding gets closer to the one-hot encoding (i.e., when the noise introduced by undertraining the CV model has smaller magnitude). In fact, our experiments show that undertraining  the CV model (as opposed to optimally training it) enhances the ability of the IE model to generalize outside the training set. 

\new{Finally, our investigation has only scratched the surface of the body of machine learning techniques that are available for learning and generalization. Alternatives to what we have considered here include probabilistic graphical models (see e.g.\ \cite{koller2009probabilistic,george2017generative}) and transformers (see e.g.\ \cite{vaswani2017attention,devlin2018bert,radford2018improving}. Whether these other methods can perform well on the identity effect tasks that are our primary examples in this paper is a worthwhile open question.}

\section*{Acknowledgments}
SB acknowledges the support of NSERC through grant RGPIN-2020-06766, the Faculty of Arts and Science of Concordia University, and the CRM Applied Math Lab. ML acknowledges the Faculty of Arts and Science of Concordia University for the financial support. PT was supported by an NSERC (Canada) Discovery Grant.

\section*{Appendix}

In this appendix we study the invariance of learning algorithms trained via the Adam method \citep{kingma2014adam} to transformations $\tau$. The setting is analogous to Section 2.1.3 of the main paper, with two main differences: (i) training is performed using the Adam method as opposed to stochastic gradient descent; (ii) the matrix $\mathcal{T}$ associated with the transformation $\tau$ is assumed to be a signed permutation matrix as opposed to an orthogonal matrix.

Consider a learning algorithm of the form $L(D,w) = f (B,Cw)$, where $D$ is our complete data set with entries $(w,r)$ and where the parameters $(B,C)$ are computed by approximately minimizing some differentiable (regularized) loss function $F(B,C) = F_{D}(B,C)$ depending on the data set $D$ (see sections 2.1.2 and 2.1.3 of the main paper).
Let $\Theta_i = (B_i,C_i)$, with $i = 0,\ldots,k$, be successive approximations obtained using the Adam method, defined by the following three update rules:
\begin{align}
\label{eq:adam_first_moment}
M^{(1)}_{i+1} & = \rho_1 M^{(1)}_i + (1-\rho_1) \frac{\partial F}{\partial \Theta} (\Theta_i), & \text{(first moments' update)}\\
\label{eq:adam_second_moment}
M^{(2)}_{i+1} & = \rho_2 M^{(2)}_i + (1-\rho_2) \left(\frac{\partial F}{\partial \Theta} (\Theta_i)\right)^{\odot 2},  & \text{(second moments' update)}\\
\label{eq:adam_update}
\Theta_{i+1} & = \Theta_i - \theta_i M^{(1)}_{i+1}\odiv \left(M^{(2)}_{i+1}\right)^{\odot \frac12}, & \text{(parameters' update)}
\end{align}
where $\odot$ and $\odiv$ are componentwise (Hadamard) product and division and $A^{\odot k}$ is the componentwise $k$th power, $0 < \rho_1, \rho_2 < 1$ are tuning parameters, and where $\frac{\partial F}{\partial \Theta} = \left(\frac{\partial F}{\partial B}, \frac{\partial F}{\partial C}\right)$. Moreover, assume $(\theta_i)_{i = 0}^{k-1}$ to be a sequence of predetermined step sizes. 

Suppose we initialize $C = C_0$ in such a way that $C_0$ and $C_0\mathcal{T}$ have the same distribution when $\mathcal{T}$ is a signed permutation. This holds, for example, when the entries of $C_0$ are identically and independently distributed according to a normal distribution $\mathcal{N}(0,\sigma^2)$.   Moreover, we initialize $B=B_0$ in some randomized or deterministic way independently of $C_0$.  The moments are initialized as $M_0^{(j)} = 0$ for $j = 1,2$.

To simplify the notation, we assume that at each step of the Adam method gradients are computed without batching, i.e.\ using the whole training data set at each iteration. We note that our results can be generalized to the case where gradients are stochastically approximated  via random batching by arguing as in Section 2.1.3 of the main paper. Moreover, we focus on the case of $\ell^2$ regularization, although a similar result holds for $\ell^1$ regularization (see Section 2.1.2 of the main paper).

Using $\ell^2$ or $\ell^1$ regularization on the parameter $C$, training the model $r = f(B,Cw)$ using the transformed data set $\tau(D)$ corresponds to minimizing the objective function $F_{\tau(D)}(B,C) = F_D(B,C\mathcal{T})$ (see Sections 2.1.2 and 2.1.3 of the main paper). We denote the sequence generated by the Adam algorithm using the transformed data set by $\Theta'_i = (B'_i,C'_i)$, with $i = 0,\ldots, k$. Now, using the chain rule
\begin{align}
\label{eq:chainrule}
\frac{\partial F_{\tau(D)}}{\partial \Theta}(\Theta'_i)
& = \left( \frac{\partial F_{\tau(D)}}{\partial B}(B'_i, C'_i),  \frac{\partial F_{\tau(D)}}{\partial C}(B'_i, C'_i)\right)\\
\nonumber
& = \left( \frac{\partial F_D}{\partial B}(B'_i, C'_i \mathcal{T} ),  \frac{\partial F_D}{\partial C}(B'_i, C'_i \mathcal{T} ) \mathcal{T}^T\right).
\end{align}
The goal is now to show that $(B'_i, C'_i \mathcal{T}) \eqdist (B_i,C_i)$ for all $i = 0,\ldots, k$ (in the sense of equidistributed random variables), so that 
$$
L(\tau(D),\tau(w)) = f(B'_k, C'_k \mathcal{T}w) \eqdist f(B_k, C_k w) = L(D,w),
$$
implying the invariance of the learning algorithm to the transformation $\tau$ corresponding to the matrix $\mathcal{T}$. This is proved in the following result.

\begin{thm} \label{thm:adam}
Let $\tau$ be a linear transformation represented by a signed permutation matrix $\mathcal{T}$.
Suppose the Adam method, as described above, is used
to determine parameters $(B_{k},C_{k})$ with the objective function
\[
F(B,C) = \mathcal{L}(f(B,Cw_i),r_i, i=1,\ldots,n) + \lambda(\mathcal{R}_1(B) +  \|C\|^2_F),
\]
for some $\lambda \geq 0$ and assume $F$ to be differentiable with respect to $B$ and $C$. Suppose the random initialization of the parameters $B$ and $C$ are independent and that the initial distribution of $C$ is invariant with respect to right-multiplication by $\mathcal{T}$.
 
Then, the learner $L$ defined by 
$L(D,w)=f(B_k,C_kw)$
 satisfies $L(D,w) \eqdist L(\tau(D),\tau(w))$. 
\end{thm}

\begin{proof}

The proof goes by induction.  We would like to show that $(B'_i, C'_i \mathcal{T}) \eqdist (B_i,C_i)$, for all $i = 0,\ldots, k$. Let $M'^{(j)}_i$ with $i = 1,\ldots,k$ and $j = 1,2$ be the sequences of first and second moments generated by the Adam method using the transformed data set. When $i = 0$, then $(B'_0, C'_0 \mathcal{T}) \eqdist (B_0,C_0)$ by assumption. Let us now assume the claim to be true for all indices less than or equal to $i$ and show its validity for the index $i+1$. 

Using the update rules \eqref{eq:adam_first_moment}--\eqref{eq:adam_update}, the chain rule \eqref{eq:chainrule} and the inductive hypothesis $(B'_i, C'_i \mathcal{T}) \eqdist (B_i,C_i)$, a direct computation shows that $B_{i+1} \eqdist B'_{i+1}$. 

Proving that $C_{i+1} \eqdist C'_{i+1}\mathcal{T}$ requires more effort. We split $M^{(j)}_i = (M^{(j,B)}_i,M^{(j,C)}_i)$ and $M'^{(j)}_i = (M'^{(j,B)}_i,M'^{(j,C)}_i)$ for $j = 1,2$. Assuming that $C_i \eqdist C'_i \mathcal{T}$ by induction and using the  update rule \eqref{eq:adam_update}, we see that
\begin{align*}
C'_{i+1} \mathcal{T} 
& = C'_i \mathcal{T}  - \theta_i \left(M'^{(1,C)}_{i+1} \odiv (M'^{(2,C)}_{i+1})^{\odot \frac12}\right) \mathcal{T} \\ 
& \eqdist C_i  - \theta_i \left(M'^{(1,C)}_{i+1} \odiv (M'^{(2,C)}_{i+1})^{\odot \frac12}\right) \mathcal{T} .
\end{align*}
Hence, a sufficient condition to have $C'_{i+1} \mathcal{T} \eqdist C_{i+1}$ is
\begin{equation}
\label{eq:needed_relation}
\left(M'^{(1,C)}_{i+1} \odiv (M'^{(2,C)}_{i+1})^{\odot \frac12}\right) \mathcal{T}  
\eqdist M^{(1,C)}_{i+1} \odiv (M^{(2,C)}_{i+1})^{\odot \frac12}.
\end{equation}
We now prove the identity \eqref{eq:needed_relation} by induction on $i$. This will in turn prove the theorem.

\paragraph{Proof of \eqref{eq:needed_relation} by induction} When $i = 0$, using the initialization $M^{(j)}_0 = M'^{(j)}_0 = 0$ for $j = 1,2$, we obtain
\begin{align*}
M^{(j,C)}_{1}  = (1-\rho_j) \left(\frac{\partial F_D}{\partial C} (B_0,C_0)\right)^{\odot j},
\qquad
M'^{(j,C)}_{1}  = (1-\rho_j) \left(\frac{\partial F_{\tau(D)}}{\partial C} (B'_0,C'_0)\right)^{\odot j},
\end{align*}
for $j = 1,2$. Therefore,
\begin{align*}
M^{(1,C)}_1 \odiv (M^{(2,C)}_1)^{\odot \frac12} 
& = \frac{1-\rho_1}{\sqrt{1-\rho_2}} \sign \left(\frac{\partial F_D}{\partial C} (B_0,C_0)\right),\\
M'^{(1,C)}_1 \odiv (M'^{(2,C)}_1)^{\odot \frac12} 
& = \frac{1-\rho_1}{\sqrt{1-\rho_2}} \sign \left(\frac{\partial F_{\tau(D)}}{\partial C} (B'_0,C'_0)\right),
\end{align*}
where $\sign(\cdot)$ is applied componentwise. Applying the chain rule \eqref{eq:chainrule}, and using that $(B'_0,C'_0 \mathcal{T}) \eqdist (B_0,C_0)$ we obtain
\begin{align*}
M'^{(1,C)}_1 \odiv (M'^{(2,C)}_1)^{\odot \frac12} 
& = \frac{1-\rho_1}{\sqrt{1-\rho_2}}\sign \left(\frac{\partial F_D}{\partial C} (B'_0,C'_0 \mathcal{T} ) \mathcal{T}^T\right)\\ 
&\eqdist \frac{1-\rho_1}{\sqrt{1-\rho_2}} \sign \left(\frac{\partial F_D}{\partial C} (B_0,C_0) \mathcal{T}^T\right).
\end{align*}
Consequently, \eqref{eq:needed_relation} holds for $i = 0$ if
$$
\sign \left(\frac{\partial F_D}{\partial C} (B_0,C_0) \mathcal{T}^T\right) \mathcal{T} \eqdist  \sign \left(\frac{\partial F_D}{\partial C} (B_0,C_0)\right).
$$
But this is true since $\mathcal{T}$ is a signed permutation matrix.

It remains to show that \eqref{eq:needed_relation} holds for $i$ assuming that it holds for all indices strictly less than $i$. To do this, we show that, for all $i = 0,\ldots, k$, we have
\begin{align}
\label{eq:aux1}
{M'}_{i}^{(1,C)} & \eqdist {M}_{i}^{(1,C)} \mathcal{T}^{\top},
\quad\text{and}
\quad
{M'}_{i}^{(2,C)} \eqdist {M}_{i}^{(2,C)} |\mathcal{T}^{\top}|,
\end{align}
where the absolute value $|\cdot|$ is applied componentwise. 

\paragraph{Proof of \eqref{eq:aux1}} The two relations in \eqref{eq:aux1} hold for $i = 0$ since $M_0^{(j)} = {M'}_0^{(j)} = 0$ for $j = 1,2$. Then, by induction,
\begin{align*}
{M'}_{i+1}^{(1,C)}
& = \rho_1 {M'}_i^{(1,C)} + (1-\rho_1) \frac{\partial F_{\tau(D)}}{\partial C}(B'_i, C'_i)\\
& \eqdist \rho_1 {M}_i^{(1,C)} \mathcal{T}^T + (1-\rho_1) \frac{\partial F_D}{\partial C}(B_i, C_i) \mathcal{T}^T\\
&= {M}_{i+1}^{(1,C)} \mathcal{T}^{\top}.
\end{align*}
Similarly, 
\begin{align*}
{M'}_{i+1}^{(2,C)}
& = \rho_1 {M'}_i^{(2,C)} + (1-\rho_1) \left(\frac{\partial F_{\tau(D)}}{\partial C}(B'_i, C'_i)\right)^{\odot 2}\\
& \eqdist \rho_1 {M}_i^{(2,C)} |\mathcal{T}^T| + (1-\rho_1) \left(\frac{\partial F_D}{\partial C}(B_i, C_i) \mathcal{T}^T\right)^{\odot 2}\\
& = \rho_1 {M}_i^{(2,C)} |\mathcal{T}^T| + (1-\rho_1) \left(\frac{\partial F_D}{\partial C}(B_i, C_i) \right)^{\odot 2}|\mathcal{T}^T|\\
& = {M}_{i+1}^{(2,C)} |\mathcal{T}^{\top}|.
\end{align*}
Thus, \eqref{eq:aux1} is valid for all $i = 0,\ldots,k$.

\paragraph{Conclusion} Finally, using \eqref{eq:aux1} and thanks to the identity $((A \mathcal{T}^T) \odiv (B |\mathcal{T}^T|)) \mathcal{T} = A \odiv B$, which is valid since $\mathcal{T}$ is a signed permutation matrix, we see that
\begin{align*}
\left({M'}^{(1,C)}_{i+1} \odiv (M'^{(2,C)}_{i+1})^{\odot \frac12}\right) \mathcal{T}  
& \eqdist \left[(M^{(1,C)}_{i+1} \mathcal{T}^T )\odiv (M^{(2,C)}_{i+1} |\mathcal{T}^T|)^{\odot \frac12}\right] \mathcal{T} \\
& = \left[(M^{(1,C)}_{i+1} \mathcal{T}^T )\odiv ((M^{(2,C)}_{i+1})^{\odot \frac12} |\mathcal{T}^T|)\right] \mathcal{T} \\
& = M^{(1,C)}_{i+1} \odiv (M^{(2,C)}_{i+1})^{\odot \frac12}.
\end{align*}
This shows \eqref{eq:needed_relation} and concludes the proof of the theorem.
\end{proof}

\bibliographystyle{apalike}
\bibliography{references}

\end{document}